\PassOptionsToPackage{table,dvipsnames}{xcolor}
\documentclass{article} 

\usepackage[margin=1in]{geometry}

\usepackage{multirow}
\usepackage{caption}
\usepackage{subcaption}
\usepackage{booktabs} 
\usepackage{mathtools}
\usepackage{amsthm}
\usepackage{enumitem}
\usepackage{wrapfig}
\usepackage{titlesec}
\usepackage{setspace}
\usepackage{svg}
\usepackage{lipsum}

\usepackage{natbib}
\usepackage{titlesec}

\definecolor{lightgray}{gray}{0.9}
\newtheorem{prop}{Proposition}

\usepackage{amsmath,amsfonts,bm}

\def\eqref#1{equation~\ref{#1}}

\def\floor#1{\lfloor #1 \rfloor}
\def\1{\bm{1}}

\def\vu{{\bm{u}}}

\def\vx{{\bm{x}}}
\def\vy{{\bm{y}}}
\def\vz{{\bm{z}}}

\DeclareMathAlphabet{\mathsfit}{\encodingdefault}{\sfdefault}{m}{sl}
\SetMathAlphabet{\mathsfit}{bold}{\encodingdefault}{\sfdefault}{bx}{n}

\def\sR{{\mathbb{R}}}
\def\sS{{\mathbb{S}}}

\usepackage{hyperref}
\hypersetup{
    colorlinks=true,
    linkcolor=blue,
    filecolor=magenta,      
    urlcolor=cyan,
    citecolor=blue
}

\title{\bf Integrating Present and Past \\in Unsupervised Continual Learning}

\author{
\stepcounter{footnote}Yipeng Zhang${}^{1,2}$\thanks{Work done partially at Columbia University.}\ , 
Laurent Charlin${}^{1,3}$,
Richard Zemel${}^{4}$, 
Mengye Ren${}^{5}$ \\\\
${}^1$Mila, ${}^2$Université de Montréal, ${}^3$HEC Montréal, ${}^4$Columbia University, ${}^5$NYU\\
\texttt{\{yipeng.zhang, lcharlin\}@mila.quebec}\\\texttt{zemel@cs.columbia.edu, mengye@cs.nyu.edu}
}
\date{}

\begin{document}

\maketitle

\begin{abstract}
We formulate a 
unifying framework for \emph{unsupervised continual learning} (UCL), which 
disentangles 
learning objectives that are specific to the present and the past data, encompassing \emph{stability}, \emph{plasticity}, and \emph{cross-task consolidation}.
The framework 
reveals that many existing UCL approaches 
overlook
cross-task consolidation and 
try to balance plasticity and stability in a shared embedding space.
This results in worse performance due to a lack of 
within-task data diversity and reduced effectiveness in learning the current task.
Our method, \texttt{Osiris}, which explicitly optimizes all three objectives on 
separate embedding spaces, 
achieves state-of-the-art performance on all benchmarks, including two novel benchmarks proposed in this paper featuring 
semantically structured task sequences. 
Compared to standard benchmarks, these two structured benchmarks more closely resemble visual signals received by humans and animals when navigating real-world environments. 
Finally, we show some preliminary evidence that continual models can benefit from such more realistic learning scenarios.
\footnote{Published as a conference paper at CoLLAs 2024. Code is available at \url{https://github.com/SkrighYZ/Osiris}.}

\end{abstract}

\section{Introduction}

Humans and animals learn visual knowledge through continuous streams of experiences. 
In machine learning, \emph{continual learning} is central to the development of learning from data that changes over time~\citep{parisi2019continual}. 
In continual learning, the learner typically encounters a non-stationary data stream through a series of learning episodes, similar to how humans learn~\citep{kurby2008segmentation}, where each episode assumes a stationary data distribution. 
When this learning process is unlabelled, it is referred to as \emph{unsupervised continual learning} (UCL)~\citep{madaan2022representational, fini2022self}. 
A popular current approach to UCL uses \emph{self-supervised learning} (SSL)~\citep{chen2020simple, zbontar2021barlow}, which aims at learning invariant representations across pairs of visually similar images. Representations learned with SSL are believed to exhibit less \emph{forgetting}~\citep{mccloskey1989catastrophic, french1999catastrophic} 
than when learned with supervised objectives such as cross-entropy~\citep{madaan2022representational, davari2022probing}.
This confers SSL an important advantage since minimizing forgetting is a central objective of continual learning~\citep{parisi2019continual}.

However, 
models trained in the UCL setting still do not perform as well as models trained from iid data (offline). 
Learning from iid data is ideal but
unrealistic when the underlying data distribution changes over time.
In contrast, in UCL, models only have access to data from the present distribution and limited access to the past.
Despite 
recent effort in advancing UCL~\citep{madaan2022representational, fini2022self, gomez2024plasticity}, limited progress has been made in closing this performance gap.

In this study, we take a step back to examine what features current UCL methods learn and why such challenges persist. Our investigation yields a unifying framework that disentangles the learning objectives specific to the present and past data. In particular, our framework jointly optimizes: \textbf{1) plasticity} for learning within the present episode, \textbf{2) consolidation} for 
integrating 
the present and the past representations, and \textbf{3) stability} for maintaining the past representations.

Our framework reveals that existing UCL methods either are not very effective at optimizing for \textbf{plasticity} or
lack an explicit formulation of \textbf{cross-task\footnote{We use \emph{task} and \emph{episode} interchangeably.} consolidation}. 
To improve plasticity, we find that it is 
crucial to 
project features to an embedding space exclusively for optimizing the current-task objective
since 
optimizing other objectives on the same space
can 
impair the model's ability to
adapt to the
new data distribution.
Meanwhile, the lack of an explicit cross-task consolidation objective reduces the data diversity within a batch and causes the present-task representations to overlap with the past ones.

We address these two limitations by 
explicitly optimizing all three objectives in our framework with two parallel projector branches. 
Our innovations are inspired by recent progress in the SSL literature on contrastive loss decomposition~\citep{wang2020understanding} and leveraging multiple embedding spaces~\citep{xiao2021what}.
We name our method \texttt{Osiris}
(\emph{\underline{O}ptimizing \underline{s}tability, plast\underline{i}city, and c\underline{r}oss-task consolidation via \underline{i}solated \underline{s}paces}). 
\texttt{Osiris} achieves state-of-the-art performance on a suite of UCL benchmarks, including the standard Split-CIFAR-100~\citep{rebuffi2017icarl} where tasks consist of randomly-drawn object classes. Additionally, we find that BatchNorm~\citep{ioffe2015batch} 
is not suitable for UCL since it presupposes a stationary distribution,
and advise future studies to use GroupNorm~\citep{wu2018group} instead.

Besides existing benchmarks, we consider the impact of structure in the sequence of episodes typically encountered in our everyday experiences.
We build temporally structured task sequences of CIFAR-100 and Tiny-ImageNet images~\citep{le2015tiny, deng2009imagenet}
that resemble
the structure of visual signals that humans and animals receive when navigating real-world environments. 
Interestingly, on the 
Structured Tiny-ImageNet benchmark, our method outperforms the offline iid model, showing some preliminary evidence that UCL algorithms can benefit from 
real-world task structures.

In summary, our main contributions are:
\begin{itemize}
    \item 
    \looseness=-100000
    We propose a unifying framework for UCL consisting of three 
    objectives
    that integrate the present and the past tasks, and show that existing methods 
    optimize a subset of these objectives but not all of them.
    
    \item We propose \texttt{Osiris}, a UCL method that directly optimizes 
    all three objectives in our framework. 

    \item For emulating more realistic learning environments, we propose two UCL benchmarks, Structured CIFAR-100 
    and Structured Tiny-ImageNet,
    that feature semantic structure on classes within or across tasks.
    We also propose 
    two new metrics to measure plasticity and consolidation
    in UCL. 
    
    \item We show that \texttt{Osiris} achieves state-of-the-art performance on all benchmarks, matching offline iid learning on the standard Split-CIFAR-100 with five tasks, and even outperforms it on the Structured Tiny-ImageNet benchmark.
\end{itemize}
\section{Preliminaries}
\label{sec:under:prelim}

\subsection{Self-Supervised Learning}
\label{sec:under:ssl}

Self-supervised learning (SSL) objectives are remarkably effective in learning good representations from unlabeled image data~\citep{chen2020simple, zbontar2021barlow, he2020momentum, caron2020unsupervised, caron2021emerging}. Their idea is to enforce the model to be invariant to low-level cropping and distortions of the image, which encourages it to encode
semantically meaningful features. We focus our analysis on the representative contrastive learning method SimCLR~\citep{chen2020simple} because it has well-studied 
geometric properties on the feature space, which can help analyses~\citep{wang2020understanding}, and it exhibits strong performance in our experiments. 

Formally, let $\mathcal{A}$ be a stochastic function that applies augmentations (random cropping, color jittering, etc.) to $\vx_i \sim \mathcal{D}$. For brevity, we fix the anchor $\vx_i$ when describing the loss and denote the two augmented views of the anchor with $\vx_i, \vx_i' = \mathcal{A}(\vx_i)$, and augmented views of other images with $\vx_j \,  (j \neq i)$. 
Let $f_\Theta: \mathcal{X} \to \sR^{d_{\text{E}}}$ 
denote our hypothesis family parameterized by $\Theta$, where $d_{\text{E}}$ is the output feature dimension of our model and $\mathcal{X} \supseteq \mathcal{D}$.
Let $g_{\Phi}: \sR^{d_{\text{E}}} \to \sR^{d_{\text{P}}}$ be a non-linear function that projects $f_\Theta(\cdot)$ to some subspace of $\sR^{d_{\text{P}}}$. Then, the contrastive loss is defined as
\begin{equation}
\label{eq:simclr}
\mathcal{L}^{\text{SSL}}(\mathcal{D}; f_\Theta, g_\Phi)
=
\mathbb{E}_{\vx_i \sim \mathcal{D}, \, \{\vx_j\}_j \overset{\mathrm{iid}}{\sim}  \mathcal{D}} \left[ -\log \frac{\exp(\vz_i^\top \vz_i') / \tau}{\exp(\vz_i^\top \vz_i') / \tau + \sum_{j\neq i} \exp(\vz_i^\top \vz_j) / \tau } \right]
\, ,
\end{equation}
where $\vz_i$ is the normalized feature vector, i.e, $\vz_i = g_\Phi(f_\Theta(\vx_i)) / \| g_\Phi(f_\Theta(\vx_i)) \|_2$
and similarly for $\vz_i'$ and $\vz_j$; 
$\vz_i, \vz_i', \vz_j \in \mathcal{S}^{d_{\text{P}} - 1}$ where $\mathcal{S}^{d_{\text{P}} - 1}$ denotes the $d_{\text{P}}$-dimensional unit sphere.
$\tau$ is a temperature hyperparameter
which we omit
for brevity 
in our analysis.
Intuitively, the gradient of this loss with respect to $\vz_i$ is a weighted (with weights in $[0, 1]$) sum of $-\vz_i'$ and every $\vz_j \, (j \neq i)$. The optimal model minimizes the distance between representations of positive pairs and maximizes the pairwise distance of different inputs.
In the remainder of this paper, we use $\mathcal{L}^{\text{SSL}}(\mathcal{D}; f_\Theta, g_\Phi)$ to denote the contrastive loss in Eq.~\ref{eq:simclr} within set $\mathcal{D}$ on the normalized output space of $g_\Phi \circ f_\Theta$ for brevity. 

\paragraph{Generalized contrastive loss.} We can extend Eq.~\ref{eq:simclr} to a more general form,
$\mathcal{L}^{\text{SSL}}(\sS_{+}, \sS_{-}; f_1, f_2)$, to denote the asymmetric contrastive loss where we use views of the same example in set $\sS_{+}$ as positive pairs and views of examples in set $\sS_{-}$ as negatives. 
One augmented view of the anchor, $\vx_i$, is encoded by $f_1$ and the comparands, $\vx_i'$ and $\vx_j$'s, are encoded by $f_2$.
Formally,
\begin{equation}
\label{eq:simclr_general}
\mathcal{L}^{\text{SSL}}(\sS_{+}, \sS_{-}; f_{1}, f_{2})
=
\mathbb{E}_{\vx_i \sim \sS_{+}, \, \{\vx_j\}_j 
\sim \sS_{-}} 
\left[ 
-\log \frac{\exp\left(s(f_{1} (\vx_i), f_{2} (\vx_i') \right)}{\exp\left(s(f_{1} (\vx_i), f_{2} (\vx_i')) \right) + \sum_{j} \exp\left(s(f_{1} (\vx_i), f_{2} (\vx_j)) \right)} 
\right]
\, ,
\end{equation}
where $s(\cdot, \cdot)$ denotes the cosine similarity function.

\subsection{Unsupervised Continual Learning}
\label{sec:under:ucl}

\sloppy
UCL studies the problem of representation learning on a set of unlabeled data distributions, $\{\mathcal{D}_1, \ldots, \mathcal{D}_T\}$, which the learner sequentially observes. Within each task $\mathcal{D}_t$, the learner is presented a batch of randomly selected examples $X = \{\vx_i\}_{i=1}^B$ 
at each step, where $\vx_i \overset{\mathrm{iid}}{\sim} \mathcal{D}_t$ and $B$ is the batch size. Suppose $\vx$ is an image that belongs to some semantic concept class; then, the learner does not know either the class label or the task label $t$ and only observes the image itself. 
The goal is to learn a good $\Theta$ such that $f_\Theta(\vx)$ encodes useful information about $\vx \sim \mathcal{X}$ that can be directly used in subsequent tasks, where $\bigcup_{t=1}^T \mathcal{D}_t \subseteq \mathcal{X}$.
A similar learning setup, with no knowledge of task labels but with class labels, 
is typically referred to as 
\emph{class-incremental learning} in the supervised continual learning (SCL) literature~\citep{vandeven2022}.

To use SSL objectives as our learning signal in UCL, the expectation in Eq.~\ref{eq:simclr} can be estimated by averaging the loss over all examples in a batch $X$. For each $\vx_i$, we use $\vx_i$ as the anchor and $X\setminus\{\vx_i\}$ as negatives. 
Two common baselines are considered in UCL:
\begin{itemize}

\item \textbf{Sequential Finetuning (FT):} At task $t$, we only sample the batch $X$ from $\mathcal{D}_t$. 

\item \textbf{Offline Training (Offline):} $X$ is sampled iid from $\mathcal{D} = \bigcup_{t=1}^T \mathcal{D}_t$ throughout training.

\end{itemize}

In SCL, \texttt{FT} serves as the performance lower bound of models trained sequentially on $\mathcal{D}_{1\ldots T}$, whereas \texttt{Offline} is expected to be a \emph{soft}\footnote{See Sec.~\ref{sec:exp:structured} for cases when UCL methods outperform \texttt{Offline}.} upper bound because it has access to the full dataset at any training step.

\section{
Dissecting the Learning Objective of UCL}

We organize this section as follows. In Sec.~\ref{sec:under:framework}, we describe 
three desirable properties for representation learning under the UCL setting. We then present \texttt{Osiris} in Sec.~\ref{sec:method}, which explicitly optimizes these properties.
Finally, in Sec.~\ref{sec:under:existing}, we show that \texttt{Osiris} is an instance of a more general framework, and that existing UCL methods implicitly address a subset of its components.

\subsection{
Three Desirable Properties
}
\label{sec:under:framework}

Features that facilitate \emph{plasticity} or \emph{stability} are commonly studied in UCL.
In this section, we highlight another category of features, which we call \emph{cross-task consolidation} features.
We argue that UCL models need to consider plasticity, stability, and consolidation, in order to achieve good performance.

\paragraph{Plasticity and stability.} Plasticity refers to the model's ability to optimize the learning objective on the \emph{present task} $\mathcal{D}_t$. 
Intuitively, \texttt{FT} usually learns the present task well because it does not consider data in other tasks.
On the other hand, stability refers to the model's ability to maintain performance on \emph{past tasks}. This is commonly achieved either by regularizing the model---with some previous checkpoints---in their parameter space~\citep{kirkpatrick2017overcoming, schwarz2018progress, chaudhry2018riemannian} or their output space~\citep{buzzega2020dark, fini2022self}, or by jointly optimizing the learning objective on some data sampled from $\mathcal{D}_{1\ldots t-1}$ so that the model still performs well in expectation on previous tasks~\citep{lin1992self, robins1995catastrophic, madaan2022representational}. In practice, the distribution $\mathcal{D}_{1\ldots t-1}$ is usually estimated online with a memory buffer.

The \emph{stability-plasticity dilemma}~\citep{ditzler2015learning} refers to the conundrum where parameters of continual learning models need to be stable in order not to forget learned knowledge but also need to be plastic to improve the representations continually. Tackling this challenge has been the main focus of prior work in both SCL and UCL.

\paragraph{Cross-task consolidation.} 
Consolidation refers to the ability to distinguish data 
from
\emph{different tasks}. 
For example, if one task contains images of cats and dogs and another contains images of tigers and wolves, then learning to contrast dogs and wolves may yield fine-grained features that help reduce cross-task errors.
Consolidation has been explored in SCL by leveraging class labels~\citep{hou2019learning, abati2020conditional, masana2021importance, kim2022theoretical} or frozen representations~\citep{aljundi2017expert, wang2023hierarchical}, but it has been overlooked in UCL. Since we want to continually improve a unified representation for all seen data without labels, existing methods are not applicable.

\subsection{Osiris: Integrating Objectives of Present and Past}
\label{sec:method}

Now we present \texttt{Osiris}, a method that explicitly optimizes plasticity, stability, and cross-task consolidation. All \texttt{Osiris}'s losses share the same encoder $f_\Theta$ but may use different nonlinear MLP projectors denoted by $g_\Phi, h_\Psi: \sR^{d_{\text{E}}} \to \sR^{d_{\text{P}}}$.
We illustrate the method in Fig.~\ref{fig:method}.

To estimate $\mathcal{D}_{1\ldots t-1}$ online, \texttt{Osiris} uses a memory buffer $\mathcal{M}$ to store data examples observed by the model so far. In this study, we assume the sampling strategy for data storing and retrieval are both uniform, with the former being achieved through online \emph{reservoir sampling}~\citep{vitter1985random}. Various works have studied non-uniform storing and retrieval~\citep{NEURIPS2019_15825aee, aljundi2019task, aljundi2019gradient, yoon2021online, gu2022not}, but they are orthogonal to this study. Throughout our analysis, we use $X$ to denote a batch of data sampled iid from $\mathcal{D}_t$ and $Y$ to denote a batch sampled iid from $\mathcal{M}$.

\subsubsection{Plasticity Loss}
\label{sec:osiris:plasticity}

The loss of the current task in the form of Eq.~\ref{eq:simclr} is 
\begin{equation}
\mathcal{L}_{\text{current}}
= 
\mathcal{L}^{\text{SSL}}\left( 
X; f_\Theta, g_\Phi
\right)
\, .
\end{equation}
It has been shown by~\cite{wang2020understanding} that, asymptotically, the perfect minimizer of $\mathcal{L}_{\text{current}}$ projects all $\vx \in \mathcal{D}_t$ uniformly to the representation space, a unit hypersphere. We hypothesize that additional losses may prevent the model from learning this solution on $\mathcal{D}_t$ effectively, 
because the uniform distribution for $\mathcal{D}_t$ is unlikely the global optima for other losses.

\looseness=-100000
Fortunately, prior SSL work offer insights on tackling the stability-plasticity dilemma: 
\emph{the backbone encoder $f_\Theta$ encodes the necessary information that helps minimize SSL losses on multiple nonlinearly projected output spaces}~\citep{xiao2021what, chen2020simple}. To learn the new task effectively, we do not apply other losses on the output space of $g_\Phi \circ f_\Theta$; additional losses are calculated on
representations 
projected from the outputs of $f_\Theta$ with some other projector $h_\Psi$. This allows the model to freely distribute $g_\Phi(f_\Theta (\vx))$ in order to optimize $\mathcal{L}_{\text{current}}$, while potentially maintaining some other distributions of $f_\Theta (\vx)$ or $h_\Psi(f_\Theta (\vx))$. The benefit of this approach is that $f_\Theta$ still encodes the features that help optimize $\mathcal{L}_{\text{current}}$ on the output space of $g_\Phi \circ f_\Theta$.

\subsubsection{Stability Loss}
\label{sec:osiris:stability}

Like most prior studies in continual learning, we introduce a loss to promote stability and reduce forgetting. We study two approaches, which we discuss next. They both use the projector $h_\Psi$.

\paragraph{Osiris-D(istillation).} 
The first approach uses distillation and requires storing a frozen checkpoint of the encoder, $f_{\Theta_{t-1}^*}$, at the end of task $t-1$. It asks the current model to predict a data example from a batch of examples 
encoded by the checkpoint, therefore encouraging it to retain previously learned features. The loss can be written with the notation of Eq.~\ref{eq:simclr_general} as 
\begin{equation}
\label{eq:past-d}
\mathcal{L}_{\text{past-D}}
= 
\frac{1}{2}\left(
\mathcal{L}^{\text{SSL}}\left( 
X, X; h_\Psi \circ f_{\Theta_{t-1}^*}, h_\Psi \circ f_\Theta
\right)
+ 
\mathcal{L}^{\text{SSL}}\left( 
X, X; h_\Psi \circ f_\Theta, h_\Psi \circ f_{\Theta_{t-1}^*}
\right)
\right)
\, .
\end{equation}
The idea of $\mathcal{L}_{\text{past-D}}$ is similar to \texttt{CaSSLe}~\citep{fini2022self}, but we distill our model with $f_{\Theta_{t-1}^*}(\vx)$ and not $g_{\Phi_{t-1}^*}(f_{\Theta_{t-1}^*}(\vx))$, where $g_{\Phi_{t-1}^*}$ is the projector checkpoint. Our approach has several advantages: (a) it has been shown that the encoder output produces better representations than the projector output~\citep{chen2020simple}; (b) the gradient of our stability loss does not pass through $g_\Phi$, which allows more freedom in 
exploring the current-task features; and (c) we do not need to store $g_{\Phi_{t-1}^*}$. In practice, we find the symmetric loss works better than having only the first term. Note that this induces negligible computational overheads because we can reuse the representations of $X$ to compute the second term.

\paragraph{Osiris-R(eplay).} The second approach does not require storing parameters.
It applies the contrastive loss on $Y$ alone.
The loss can be written as  
\begin{equation}
\mathcal{L}_{\text{past-R}}
= 
\mathcal{L}^{\text{SSL}}\left( 
Y; f_\Theta, h_\Psi
\right)  
\, .    
\end{equation}
This loss prevents forgetting by optimizing the learning objective on $\mathcal{D}_{1\ldots t-1}$, in expectation. It is similar to \texttt{ER},
to be discussed in Sec.~\ref{sec:under:existing},
but uses a different projector.

\paragraph{Remark.}
Although using both $\mathcal{L}_{\text{past-R}}$ and $\mathcal{L}_{\text{past-D}}$ may yield better performance, we emphasize that this is not our goal. Because both losses aim to reduce forgetting, we use them to demonstrate the flexibility of our framework and investigate the pros and cons of replay versus distillation in our experiments. 
We expect that it is possible
to use any replay or output regularization method explored in SCL (e.g., \texttt{DER} in~\citealt{buzzega2020dark}) as our stability loss, as long as they operate on the output space of $h_\Psi$.

\begin{figure}[t]
\begin{center}
\includegraphics[width=1\linewidth]{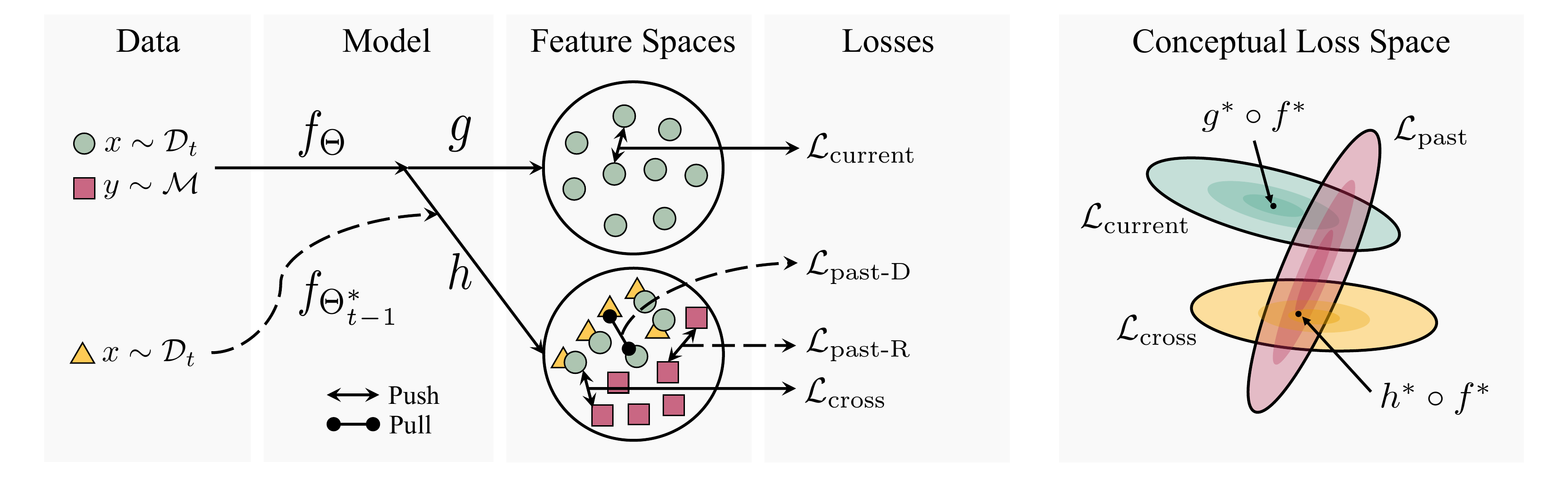}
\end{center}
\caption{Left: illustration of our method. Dashed arrows denote optional computations because the stability loss $\mathcal{L}_{\text{past}}$ can be achieved through distillation or replay. Right: conceptual loss space. A separate projector helps with optimization.}
\label{fig:method}
\end{figure}

\subsubsection{Cross-Task Consolidation Loss}
\label{sec:osiris:consolidation}

Recall from Sec.~\ref{sec:osiris:plasticity} that the perfect minimizer of $\mathcal{L}_{\text{current}}$ projects all data from $\mathcal{D}_t$ uniformly to the representation space. 
Similarly, representations of $\mathcal{D}_{1\ldots t-1}$ encoded by the perfect minimizer of $\mathcal{L}_{\text{past}}$ are also distributed uniformly.
This means 
that the model may still suffer from representation overlaps between $\mathcal{D}_{1\ldots t-1}$ and $\mathcal{D}_t$ (or between a pair of tasks from $\mathcal{D}_{1\ldots t-1}$) even if they successfully optimize $\mathcal{L}_{\text{current}}$ and $\mathcal{L}_{\text{past}}$.
Although using separate projectors for $\mathcal{L}_{\text{current}}$ and $\mathcal{L}_{\text{past}}$ may help, we propose to introduce a loss that explicitly reduces the overlap.

Consider features that are useful in discriminating instances of $\mathcal{D}_t$ from those of $\mathcal{D}_{1\ldots t-1}$; they may not be readily encoded in $f_{\Theta_{t-1}^*}$, as the model has not seen any data from $\mathcal{D}_t$ at the end of task $t-1$. Thus, distillation does not help much in this case. Instead, we propose to leverage the memory $\mathcal{M}$. 
We find using an additional projector for this loss is unnecessary, so we reuse the output space of $h_\Psi$. 
Our consolidation loss is
\begin{equation}
\mathcal{L}_{\text{cross}}
= 
\frac{1}{2}
\left(\mathcal{L}^{\text{SSL}}\left( 
X, Y; h_\Psi \circ f_\Theta, h_\Psi \circ f_\Theta
\right) + 
\mathcal{L}^{\text{SSL}}\left( 
Y, X; h_\Psi \circ f_\Theta, h_\Psi \circ f_\Theta
\right)\right) \, .
\end{equation}
This loss contrasts the current task and the memory, which promotes learning features that help discriminate the instances from the current task and past tasks. 
Similar to Eq.~\ref{eq:past-d}, we use a symmetric loss for $\mathcal{L}_{\text{cross}}$.

\paragraph{Remark.}
One might expect that $\mathcal{L}_{\text{cross}}$ encourages representations of $\mathcal{M}$ to collapse to a single point to minimize their similarity with $\mathcal{D}_t$.
We believe that this is unlikely because we find empirically that the stability loss helps the model to learn well-behaved representations of $\mathcal{M}$.
For example, optimizing $\mathcal{L}_{\text{past-D}}$, which is calculated on $\mathcal{D}_t$, yields features that are transferable
to $\mathcal{M}$.
Similarly, $\mathcal{L}_{\text{past-R}}$ directly prevents collapses with the contrastive loss on $\mathcal{M}$.
In addition, since data storing is performed online, $\mathcal{M}$ may contain examples from $\mathcal{D}_t$ while learning it, causing $\mathcal{L}_{\text{cross}}$ to contrast examples within $\mathcal{D}_t$. Nevertheless, because most SSL objectives encourage discrimination on the instance level rather than the class level, the model would still learn useful features with $\mathcal{L}_{\text{cross}}$ in this scenario.

\subsubsection{Overall Loss}
The overall loss of our model is
\begin{equation}
\mathcal{L} = \mathcal{L}_{\text{current}} + \frac{1}{2}\left( \mathcal{L}_{\text{cross}} + \mathcal{L}_{\text{past}}\right) \, .
\end{equation}
Similarly to~\cite{fini2022self}, we do not perform hyperparameter tuning on the loss weights (although it might yield better results) and fix the additional weights to sum to one to demonstrate the potential of this framework. 
In Fig.~\ref{fig:method}, we illustrate the conceptual optimization landscape. 
As discussed above, no solution minimizes all three losses individually in the same space. 
With isolated features spaces, our method reduces the extent to which $\mathcal{L}_{\text{cross}}$ or $\mathcal{L}_{\text{past}}$ directly constrain the model 
from learning
the current task, which promotes plasticity.  
Note that $f_\Theta$ still needs to maintain a unified representation that preserves information useful for minimizing all three losses.

\subsection{Expressing UCL Methods with A Unifying Framework}
\label{sec:under:existing}

Now, we extend \texttt{Osiris} to a more general framework in the form of a unified optimization objective, based on the encoder-projector architecture of SSL models. 
At task $t$, it can be expressed as the following:
\begin{multline}
\label{eq:framework}
\mathcal{L}_{X \overset{\mathrm{iid}}{\sim} \mathcal{D}_t, Y \overset{\mathrm{iid}}{\sim} \mathcal{M}}^* \left( X, Y; f_\Theta, f_{\Theta_{t-1}^*}, g_{\Phi}, g_{\Phi_{t-1}^*}, h_{\Psi}, m_{\Omega} \right) \\
= \mathcal{L}_{\text{current}}(X; f_\Theta, g_{\Phi})
+ \lambda_1\mathcal{L}_{\text{cross}}(X, Y; f_\Theta, h_{\Psi})
+ \lambda_2\mathcal{L}_{\text{past}}(X, Y; f_{\Theta}, f_{\Theta_{t-1}^*}, g_{\Phi}, g_{\Phi_{t-1}^*}, m_{\Omega})
\, .
\end{multline}
Here, 
we reuse the notations from Sec.~\ref{sec:method} and introduce some new ones:
$m_{\Omega}$ is an additional nonlinear MLP projector parameterized by $\Omega$, and  
$\lambda_1$,$\lambda_2$ denote the loss weights ($\lambda_1, \lambda_2 \geq 0$). 
The three terms in Eq.~\ref{eq:framework} learn features that promote plasticity, cross-task consolidation, and stability, respectively.
Note that it is not necessary 
for the objective
to use all the arguments included in the parentheses
or 
different parameterizations for $g_\Phi, h_\Psi, m_\Omega$;
we only use Eq.~\ref{eq:framework} as the general form. 
\emph{Nevertheless, an ideal model explores good features from the first two terms and uses the third term to ensure that it does not forget them over time.}

Interestingly, most existing UCL methods implicitly optimize the terms in Eq.~\ref{eq:framework}, which we formally show next.
Our analysis 
is similar to~\cite{wang2024a}, but is distinct in that we decompose the objective from a representation learning perspective rather than a methodology perspective. We fix the first term (shared across methods) and let $f_\Theta$ and $g_\Phi$ be the 
current-task encoder and projector as described in Sec.~\ref{sec:under:ssl}. One exception is dynamic model architectures where new parameters are added during learning~\citep{yoon2018lifelong, rusu2016progressive}; we do not discuss architecture-based methods as it would be equivalent to progressively adding arguments to the first term. 
We illustrate the focus of different UCL methods in Table~\ref{tab:framework}.

\paragraph{Elastic Weight Consolidation~(EWC)} is a classic baseline in continual learning~\citep{kirkpatrick2017overcoming, schwarz2018progress, chaudhry2018riemannian}. It uses $(\Theta - \Theta_{t-1}^*)^\top F_t (\Theta - \Theta_{t-1}^* )$ as $\mathcal{L}_{\text{past}}$, where $F_t$ is the diagonal Fisher information matrix at task $t$ that can be estimated with $F_{t-1}, \mathcal{D}_{t-1}, f_{\Theta_{t-1}^*}$, and $g_{\Phi_{t-1}^*}$ at the end of task $t-1$. \texttt{EWC} does not use a memory buffer, but needs to store $F_t$ and $\Theta_{t-1}^*$.  
It sets $\lambda_1 = 0$ and does not consider cross-task consolidation.

\begin{table}[t]
\caption{Comparison of UCL methods based on the feature components optimized. 
$^\ddagger$Analysis for \texttt{LUMP} is based on applying \emph{mixup} in the latent space.
}
\label{tab:framework}
\begin{center}
\resizebox{0.7\linewidth}{!}{
\renewcommand{\arraystretch}{1.2}
\begin{tabular}{ll|ccc}
\toprule
& {\bf Method} & Isolated Space for $\mathcal{L}_{\text{current}}$ & $\mathcal{L}_{\text{cross}}$ & $\mathcal{L}_{\text{past}}$ \\

\midrule
&FT & $\checkmark$ & & \\

\midrule

\parbox[t]{3mm}{\multirow{5}{*}{\rotatebox[origin=c]{90}{\textsc{Replay}}}}
&ER~\citep{lin1992self, robins1995catastrophic}  & & & $\checkmark$ \\

&ER+ & & $\checkmark$& $\checkmark$ \\
&ER++ & & $\checkmark$& $\checkmark$ \\

&DER~\citep{buzzega2020dark} & & & $\checkmark$ \\

&LUMP$^\ddagger$~\citep{madaan2022representational} & & $\checkmark$ & $\checkmark$ \\

\midrule

\parbox[t]{3mm}{\multirow{3}{*}{\rotatebox[origin=c]{90}{\textsc{Distill.}}}}

&EWC~\citep{schwarz2018progress} & & & $\checkmark$ \\

&CaSSLe~\citep{fini2022self} & $\checkmark$ & & $\checkmark$ \\

&POCON~\citep{gomez2024plasticity} & $\checkmark$ & & $\checkmark$ \\

\midrule
\parbox[t]{3mm}{\multirow{2}{*}{\rotatebox[origin=c]{90}{\textsc{Ours}}}}
&Osiris-R (Replay) & $\checkmark$ & $\checkmark$ & $\checkmark$ \\
&Osiris-D (Replay+Distillation) & $\checkmark$ & $\checkmark$ & $\checkmark$ \\
\bottomrule
\end{tabular}
}
\end{center}
\end{table}

\paragraph{CaSSLe}\citep{fini2022self} is the previous state of the art in UCL. 
It uses the SSL objective as $\mathcal{L}_{\text{past}}$ to regularize the model on a separate output space projected from the main model output. The \texttt{CaSSLe} loss can be expressed 
with Eq.~\ref{eq:simclr_general} as $\mathcal{L}^{\text{SSL}}(X, X; m_\Omega \circ f_\Theta, g_{\Phi_{t-1}^*} \circ f_{\Theta_{t-1}^*})$.
\texttt{CaSSLe} uses another projector $h$ on the output of $g$ to form $m$, i.e., $m_\Omega = h_\Psi \circ g_\Phi$. 
This objective encourages the main model $g_\Phi \circ f_\Theta$ to encode information that can be used to predict representations of 
a previous checkpoint,
$g_{\Phi_{t-1}^*} \circ f_{\Theta_{t-1}^*}$, thereby restricting the main model from losing features learned.
The current task loss still acts on $g_\Phi \circ f_\Theta$. Since the gradient from the regularization is back-propagated to $g$, \texttt{CaSSLe} may still limit the model's ability to learn the new task effectively. It also does not consider cross-task consolidation ($\lambda_1 = 0$) and does not use a memory buffer.

\paragraph{Experience Replay~(ER)}\citep{lin1992self, robins1995catastrophic} is a classic replay-based baseline. It uses $\mathcal{L}^{\text{SSL}}(Y; f_\Theta, g_\Phi)$ as $\mathcal{L}_{\text{past}}$ and sets $\lambda_1 = 0$. Unlike regularization-based methods, this loss may implicitly allow the model to discover new features that improve cross-task consolidation since $Y$ includes examples from $\mathcal{D}_{1\ldots t-1}$.

\paragraph{ER+ and ER++} 
\looseness=-100000
are our attempts to improve \texttt{ER}
which,
in addition to \texttt{ER},
exploit the abundance of the current task data for $\mathcal{L}_{\text{cross}}$.
One such way is to add the memory examples into negatives augmenting the current task negatives, i.e., 
$\mathcal{L}^{\text{ER+}} = \mathcal{L}^{\text{SSL}}(X, X \cup Y; g_\Phi \circ f_\Theta, g_\Phi \circ f_\Theta)$; it is similar to the asymmetric loss used by~\cite{cha2021co2l} in SCL and we refer to it as \texttt{ER+}. This loss indeed pushes 
representations of the current task and the memory
apart, but it does not yield gradient that enforces \emph{alignment}~\citep{wang2020understanding} of different views generated from memory examples. An alternative way is to use a full SSL loss on the union of the current batch and memory, i.e., $\mathcal{L}^{\text{ER++}} = \mathcal{L}^{\text{SSL}}(X \cup Y; f_\Theta, g_\Phi)$. We refer to this method as \texttt{ER++}.

\paragraph{Dark Experience Replay~(DER)}\citep{buzzega2020dark, madaan2022representational} is an improved version of \texttt{ER}, 
where $\lambda_1$ remains $0$ and $\mathcal{L}_{\text{past}}$ becomes a regularizer on the output space which is empirically estimated with $\frac{1}{|Y|}\sum_{\vy \in Y}\|g_\Phi(f_\Theta(\vy)) - \vz\|_2^2$, where $\vz$ is the representation $g(f(\vy))$ encoded by the model when $\vy$ is stored into $\mathcal{M}$. It does not consider cross-task consolidation because this objective does not encourage learning new features.

\paragraph{LUMP}\citep{madaan2022representational} is the state-of-the-art replay-based UCL method. It applies \emph{mixup}~\citep{zhang2018mixup}, a linear interpolation between $\vx\in X$ and $\vy\in Y$, to generate inputs: $\Tilde{\vx}_i = \nu \vx_i + (1-\nu) \vy_i$ where $\nu \sim \text{Beta}(\alpha, \alpha)$ with $\alpha$ being a hyperparameter. The batch $\Tilde{X} = \{\Tilde{\vx_i}\}_{i=1}^B$ is passed to the only loss term of the model, $\mathcal{L}^{\text{SSL}}(\Tilde{X}; f_\Theta, g_\Phi)$. This loss form does not fall directly into our framework because it is hard to disentangle the effect of the loss on points in between data examples. We give the proposition below and prove it in Appendix~\ref{sec:appendix:lump_linearity}.

\begin{prop}
\label{prop:lump}
\sloppy
Let $\nu \sim \text{Beta}(\alpha, \alpha)$ and let $\mathcal{L}^{\text{LUMP}}(X, Y; \nu, f_\Theta, g_\Phi) \coloneqq \mathcal{L}^{\text{SSL}}(\Tilde{X}; f_\Theta, g_\Phi)$ be as described above.
Define $\vz_i \coloneqq g(f(\vx_i)) / \|g(f(\vx_i))\|_2$, $\vu_i \coloneqq g(f(\vy_i)) / \|g(f(\vy_i))\|_2$, and $\Tilde{\vz_i} \coloneqq g(f(\Tilde{\vx_i})) / \|g(f(\Tilde{\vx_i}))\|_2$ for all $i \in \{1, \ldots |X|\}$. Suppose that the representations are linear in between $\vx_i$ and $\vy_i$, i.e., $\Tilde{\vz_i} = \nu \vz_i + (1-\nu) \vu_i$. Then
\begin{align}
\mathbb{E}_\nu \left[ \frac{\partial \mathcal{L}^{\text{LUMP}}}{\partial \vz_i} \right] 
&=
\underbrace{-a_i \vz_i' + \sum_{j \neq i} a_j \vz_j}_{\mathcal{L}_{\text{current}}} + \underbrace{\sum_{j \neq i} b_j \vu_j}_{\mathcal{L}_{\text{cross}}} - b_i \vu_i'
\, , \\
\text{and} \quad
\mathbb{E}_\nu \left[ \frac{\partial \mathcal{L}^{\text{LUMP}}}{\partial \vu_i} \right] 
&=
\underbrace{- c_i \vu_i' + \sum_{j \neq i} c_j \vu_j}_{\mathcal{L}_{\text{past}}} + \underbrace{\sum_{j \neq i} d_j \vz_j}_{\mathcal{L}_{\text{cross}}}
- d_i \vz_i'
\, ,
\end{align}
where $a_{(\cdot)}, b_{(\cdot)}, c_{(\cdot)}, d_{(\cdot)} \geq 0$ are scalar functions of $\alpha$ and the softmax probabilities of predictions. 
\end{prop}

The first equation represents the gradient of the loss w.r.t. representations of examples in the current-task batch $X$. In contrast, the second equation represents the gradient w.r.t. examples in the batch $Y$ sampled from the memory. 
Recall from Sec.~\ref{sec:under:ssl} that $(\cdot)'$ denotes the representation of another view of the same input. The linearity assumption we make above may not seem to be true in general, but note that \emph{mixup} aims to help the model behave linearly in between examples to generalize better~\citep{zhang2018mixup}. 
Nevertheless, we aim to estimate \texttt{LUMP}'s effect with this decomposition. Besides the last term in each equation, which indicates the gradient that pulls $\vz_i$ and $\vu_i$ (the pair of examples being mixed) closer, Prop.~\ref{prop:lump} says that the gradient of \texttt{LUMP} with contrastive learning can be decomposed and affects all components of Eq.~\ref{eq:framework}:
(a) current task learning, (b) cross-task discrimination, and (c) past task learning.
However, since all the losses act on the same output space and the coefficients are correlated, this does not allow flexible control of the learning emphasis.

\section{Experiments}
\label{sec:exp}

We organize this section as follows.
In Sec~\ref{sec:exp:protocol}, we describe our baselines and benchmarks, and discuss the incompatibility between BatchNorm and UCL. 
We then detail the evaluation metrics used in this study in Sec.~\ref{sec:exp:eval}.
We show our results on the standard benchmarks in Sec.~\ref{sec:exp:standard} and on the structured benchmarks in Sec.~\ref{sec:exp:structured}. 
Finally, in Sec.~\ref{sec:exp:analysis}, we analyze the models' behavior with additional experiments.

\subsection{Experimental Protocol}
\label{sec:exp:protocol}

\paragraph{Baselines.} 
We compare our method with \texttt{FT} and \texttt{Offline} defined in Sec.~\ref{sec:under:ucl}. We also compare with classic CL methods: online \texttt{EWC}~\citep{schwarz2018progress}, \texttt{ER}~\citep{lin1992self, robins1995catastrophic}, \texttt{DER}~\citep{buzzega2020dark} which we described in Sec.~\ref{sec:under:existing}. We find that we can improve \texttt{DER}'s performance by normalizing the features before performing L2 regularization, so we report the performance of improved \texttt{DER} only. We also report the performance of \texttt{ER+} and \texttt{ER++} described
in Sec.~\ref{sec:under:existing}. In addition to these standard continual learning baselines, we compare with methods designed specifically for UCL, 
\texttt{CaSSLe}~\citep{fini2022self} and \texttt{LUMP}~\citep{madaan2022representational}. Concurrent work, \texttt{POCON}~\citep{gomez2024plasticity}, also aims to maximize plasticity for UCL. We use the online version of \texttt{POCON} for a fair comparison so that all methods observe the data the same number of times. 
Finally, we do not compare with \texttt{C$^2$ASR}~\citep{cheng2023contrastive} because it requires sorting the entire task stream and is an additional plug-in for UCL methods. 

\begin{figure}[t]
\begin{center}
\includegraphics[width=0.6\linewidth]{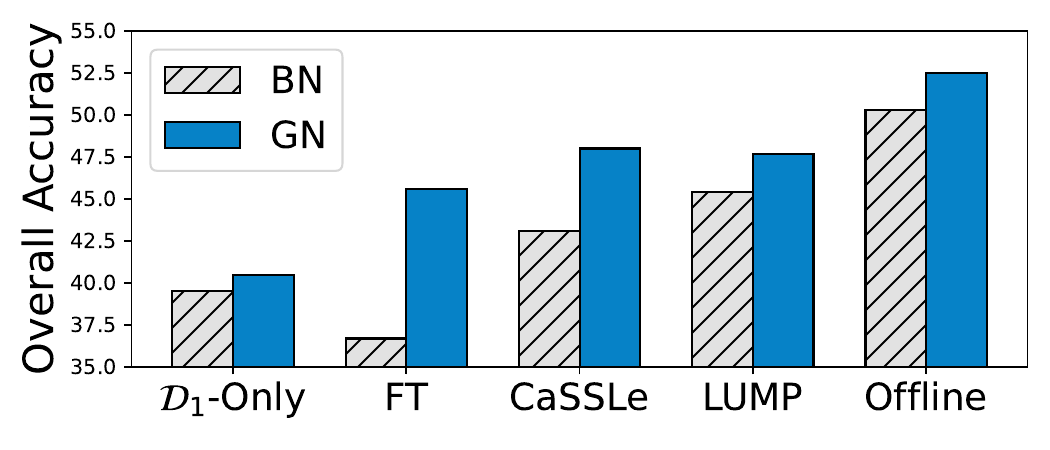}
\end{center}
\caption{
KNN accuracy of methods trained on the 20-task Split-CIFAR-100 with BatchNorm (BN) or GroupNorm (GN). $\mathcal{D}_1$\texttt{-Only} denotes an offline model trained for the same number of steps but only on the first task. \texttt{CaSSLe}~\citep{fini2022self} and \texttt{LUMP}~\citep{madaan2022representational} are state-of-the-art UCL methods. The incompatibility between BN and UCL can be mitigated by using GN instead. 
}
\label{fig:exp:norm}
\end{figure}

\paragraph{Benchmarks.} 
\begin{itemize}[leftmargin=*]
    \item \textbf{Standard Split-CIFAR-100.} Following~\cite{madaan2022representational, fini2022self}, we evaluate the models on the 5-task and 20-task sequences of CIFAR-100~\citep{krizhevsky2009learning}. It contains 50,000 32$\times$32 images from 100 classes that are randomly grouped into a disjoint set of tasks.
    
    \item \textbf{Structured CIFAR-100.} 
    In real-world environments, consecutive visual scenes are often similar and correlated in time. For example, 
    seeing a great white shark immediately followed by office chairs is rather unlikely.
    We construct a temporally structured CIFAR-100 sequence by grouping
    the classes with the same superclass label (provided by the dataset) into a task and randomly shuffle the task order, which results in ten tasks. Examples of superclasses include vehicles, flowers, and aquatic mammals.
    
    \item \textbf{Structured Tiny-ImageNet.} 
    Real-world environments also boast an abundance of hierarchies. We may visit different city blocks in an urban area and then tour multiple spots in a wild park. To create a task sequence that captures hierarchical environment structure, 
    we use Tiny-ImageNet-200~\citep{le2015tiny, deng2009imagenet}, 
    which includes 100,000 images of size 64$\times$64 categorized into 200 classes whose location spans different environments. We first use a pre-trained scene classifier trained on Places365~\citep{zhou2017places} to classify all images into \texttt{indoor}, \texttt{city}, and \texttt{wild} environments. We then use a majority vote to decide the environment label for each class. Finally, we arrange the classes in the order of \texttt{indoor} $\to$ \texttt{city} $\to$ \texttt{wild} and group them into ten tasks in order. This leads to four tasks indoors, three tasks in the city, and three tasks in the wild.
    Compared with Structured CIFAR-100, which only enforces correlation within each task, this benchmark additionally imposes correlation between consecutive tasks.
    It provides a realistic structure,
    but also aligns nicely with the classic task-incremental learning setup. 
\end{itemize}

\paragraph{Implementation details.} We use a single-head ResNet-18~\citep{he2016deep} as our backbone encoder. 
We use a two-layer MLP with a hidden dimension of 2048 and an output dimension of 128 as the projector. We use the ReLU function as activation after the hidden layer but not the output layer following~\cite{chen2020simple}. For \texttt{POCON}, we follow the authors' implementation and use four-layer projectors for distillation. We set the memory size $|\mathcal{M}|=500$ for all experiments. We train the models with a batch size of 256 for 200 epochs for UCL training, 
following~\cite{madaan2022representational}. All methods use the same loss as \texttt{FT} during the first task. 
We provide additional training, data augmentation hyperparameters, and other details in Appendix~\ref{sec:appendix:implementation}.

\paragraph{Incompatibility between BatchNorm and UCL.}

It has been shown that BatchNorm~\citep{ioffe2015batch} is not suitable for SCL since its running estimates of the feature moments (over the batch dimension) are biased towards the most recent task~\citep{pham2021continual}.
An alternative normalization layer is GroupNorm~\citep{wu2018group}, where the batch statistics are not needed 
because the normalization is applied along the feature dimension.
The performance of BatchNorm and GroupNorm has not been investigated in UCL, 
although UCL usually requires much more training iterations than SCL.
Thus, we hypothesize that using BatchNorm is harmful in UCL and the improvement of UCL methods over \texttt{FT} may not be as large as previously believed.
In Fig.~\ref{fig:exp:norm}, we show that after we switch to GroupNorm, \texttt{FT} becomes a very strong baseline, and existing UCL methods do not help as much,
although all methods show improved performance. Moreover, a model trained on only one task outperforms \texttt{FT} when equipped with BatchNorm but not GroupNorm, further showing the detrimental effect of BatchNorm in UCL. Therefore, we use GroupNorm in our experiments to 
highlight the core factors contributing to UCL performance.

\subsection{Metrics for Evaluating UCL Methods}
\label{sec:exp:eval}
We have discussed the elements contributing to learning a good representation in UCL. We now introduce accompanying fine-grained metrics to evaluate them. After unsupervised training, we keep the encoders and discard the projectors following standard practice because projectors learn the invariance that minimizes the SSL loss and may discard too much information for downstream tasks~\citep{chen2020simple}. Following~\cite{madaan2022representational}, let $A_{t, i}$ be the weighted KNN~\citep{wu2018unsupervised} test accuracy of the encoder $f_{\Theta_t^*}$ on task $i$ after it finishes training on task $t$. Note that the KNN classifier does not know the task labels (see Appendix~\ref{sec:appendix:wtl} for results when task labels are given); therefore, it is 
important for $f_{\Theta_t^*}$ to obtain a good representation geometry over the entire dataset. We report the mean and standard deviation of results obtained from three random seeds in all of our tables and plots.
We use five metrics in this study; accuracy, forgetting, and forward transfer are commonly used and we propose knowledge gain and cross-task consolidation score to measure the model's ability to optimize $\mathcal{L}_{\text{current}}$ and $\mathcal{L}_{\text{cross}}$ in 
Eq.~\ref{eq:framework}:

\begin{itemize}[leftmargin=*]
    \item \textbf{Overall Accuracy (A)} is the accuracy of the final model on all classes in the dataset:
    $A = \frac{1}{T}\sum_{i=1}^T A_{T, i}$. 

    \item \textbf{Forgetting (F)} 
    measures the difference between the model's best accuracy on task $i$ at any point during training and its accuracy on task $i$ after training:
    $F = \frac{1}{T-1}\sum_{i=1}^{T-1} \max_{t \in \{1, \ldots T\}} (A_{t, i} - A_{T, i})$. Forgetting measures the model's ability to optimize $\mathcal{L}_{\text{past}}$ in Eq.~\ref{eq:framework}. 

    \item \textbf{Knowldege Gain (K).} It has been shown that representations learned in UCL are significantly less prone to forgetting and are more stable than their SCL counterparts~\citep{davari2022probing, madaan2022representational}. On the other hand, models learned continually lose plasticity~\citep{pmlr-v232-abbas23a}. 
    Thus, we use knowledge gain to quantify the accuracy increase on task $i$ before and after the model is trained on it. It is defined as $K = \frac{1}{T-1}\sum_{i=2}^{T} (A_{i, i} - A_{i-1, i})$. Knowledge gain is similar to the SCL
    metrics proposed by~\cite{chaudhry2018riemannian, koh2023online}, 
    but it is simple to calculate and is more suitable for UCL because UCL models 
    generally have a reasonable performance on a task before even learning on it
    thanks to the generalization capability of SSL. Knowledge gain quantifies the model's ability to optimize $\mathcal{L}_{\text{current}}$. 

    \item \textbf{Cross-Task Consolidation (C)} is defined 
    as the test accuracy of a task-level KNN classifier on the frozen representations of the final model. As discussed in the previous sections, new knowledge acquisition is quantified by both knowledge gain and the ability to learn features that discriminate data across tasks, i.e., to optimize $\mathcal{L}_{\text{cross}}$. 

    \item \textbf{Forward Transfer (T)} quantifies the generalization ability of UCL models by measuring how much of the learned representation can be helpful to an unseen task. It is defined as $T = \frac{1}{T-1}\sum_{i=2}^{T} (A_{i-1, i} - R_i)$ where $R_i$ is the accuracy of a randomly initialized model on task $i$. It is used by~\cite{madaan2022representational, fini2022self}.
\end{itemize}

\begin{table}[t]
\caption{Results on standard Split-CIFAR-100 with five or 20 tasks. The best model in each column is made \textbf{bold} and the second-best model is \underline{underlined}. $^\dagger$Improved \texttt{DER}. $^{\mathsection}$Online version of \texttt{POCON}. We separate replay-based (top) and distillation-based methods (bottom) for easier comparisons.}
\label{tab:random}
\begin{center}
\resizebox{1\linewidth}{!}{
\setlength{\tabcolsep}{4pt}
\begin{tabular}{l|ccccc|ccccc}
\toprule
& \multicolumn{5}{c|}{\textsc{5-Task Split-CIFAR-100}} & \multicolumn{5}{c}{\textsc{20-Task Split-CIFAR-100}}\\

& 
\bf A ($\uparrow$) & \bf F ($\downarrow$) & \bf K ($\uparrow$) & \bf C ($\uparrow$) & \bf T ($\uparrow$) & 
\bf A ($\uparrow$) & \bf F ($\downarrow$) & \bf K ($\uparrow$) & \bf C ($\uparrow$) & \bf T ($\uparrow$) \\

\midrule

FT & 50.7 \tiny{($\pm$ 0.4)} & 2.9 \tiny{($\pm$ 0.0)} & 9.0 \tiny{($\pm$ 0.1)} & 59.1 \tiny{($\pm$ 0.2)} & 30.7 \tiny{($\pm$ 0.3)} & 45.6 \tiny{($\pm$ 0.3)} & 2.9 \tiny{($\pm$ 0.4)} & 2.5 \tiny{($\pm$ 0.1)} & 47.2 \tiny{($\pm$ 0.1)} & 28.9 \tiny{($\pm$ 0.2)} \\

\midrule

ER
& 51.5 \tiny{($\pm$ 0.4)} & 2.7 \tiny{($\pm$ 0.4)} & 8.4 \tiny{($\pm$ 0.3)} & 59.7 \tiny{($\pm$ 0.4)} &  \underline{31.8} \tiny{($\pm$ 0.2)} & 47.1 \tiny{($\pm$ 0.7)} & 3.4 \tiny{($\pm$ 0.6)} & 3.5 \tiny{($\pm$ 0.4)} & 48.2 \tiny{($\pm$ 0.5)} & 30.1 \tiny{($\pm$ 0.2)} \\

DER$^\dagger$
& 51.0 \tiny{($\pm$ 0.6)} & 3.0 \tiny{($\pm$ 0.7)} & 9.6 \tiny{($\pm$ 0.3)} & 59.0 \tiny{($\pm$ 0.4)} & 30.5 \tiny{($\pm$ 0.1)} & 45.7 \tiny{($\pm$ 0.1)} & 2.6 \tiny{($\pm$ 0.2)} & 2.6 \tiny{($\pm$ 0.4)} & 47.2 \tiny{($\pm$ 0.2)} & 28.7 \tiny{($\pm$ 0.2)} \\

LUMP
& 50.2 \tiny{($\pm$ 0.6)} & \underline{1.4} \tiny{($\pm$ 1.1)} & 7.3 \tiny{($\pm$ 0.3)} & 58.4 \tiny{($\pm$ 0.4)} & 30.2 \tiny{($\pm$ 0.1)} & 47.7 \tiny{($\pm$ 1.1)} & 2.6 \tiny{($\pm$ 0.9)} & 3.1 \tiny{($\pm$ 0.3)} & 49.1 \tiny{($\pm$ 1.0)} & 29.7 \tiny{($\pm$ 0.0)} \\

ER+ & 51.8 \tiny{($\pm$ 0.6)}& 3.4 \tiny{($\pm$ 0.5)} & {\bf 10.2} \tiny{($\pm$ 0.3)} & \underline{60.1} \tiny{($\pm$ 0.3)} & 31.1 \tiny{($\pm$ 0.4)} & 46.7 \tiny{($\pm$ 0.3)} & 3.1 \tiny{($\pm$ 0.3)}& 4.4 \tiny{($\pm$ 0.0)} & 48.0 \tiny{($\pm$ 0.1)} & 28.8 \tiny{($\pm$ 0.1)} \\

ER++ & 51.8 \tiny{($\pm$ 0.3)} & 2.9 \tiny{($\pm$ 0.6)} & 9.1 \tiny{($\pm$ 0.4)} & 59.8 \tiny{($\pm$ 0.3)} & 31.6 \tiny{($\pm$ 0.4)} & 47.7 \tiny{($\pm$ 0.3)} & 3.7 \tiny{($\pm$ 0.3)} & {\bf 5.0} \tiny{($\pm$ 0.4)} & 49.0 \tiny{($\pm$ 0.5)} & 30.0 \tiny{($\pm$ 0.1)} \\

\rowcolor{lightgray} Osiris-R (Ours) & \underline{52.3} \tiny{($\pm$ 0.5)} & 2.5 \tiny{($\pm$ 0.7)} & 8.5 \tiny{($\pm$ 0.1)} & \underline{60.1} \tiny{($\pm$ 0.1)} & {\bf 32.1} \tiny{($\pm$ 0.2)} & \underline{49.3} \tiny{($\pm$ 0.3)} & 3.1 \tiny{($\pm$ 0.2)} & \underline{4.7} \tiny{($\pm$ 0.2)} & \underline{50.5} \tiny{($\pm$ 0.5)} & {\bf 31.5} \tiny{($\pm$ 0.3)} \\

\midrule

EWC
& 43.8 \tiny{($\pm$ 0.6)} & 2.3 \tiny{($\pm$ 0.8)} & 5.0 \tiny{($\pm$ 0.4)} & 53.9 \tiny{($\pm$ 0.7)} & 26.4 \tiny{($\pm$ 0.3)} & 37.6 \tiny{($\pm$ 0.2)} & {\bf 1.7} \tiny{($\pm$ 0.1)} & 2.0 \tiny{($\pm$ 0.3)} & 39.4 \tiny{($\pm$ 0.3)} & 21.4 \tiny{($\pm$ 0.4)} \\

CaSSLe
& 51.2 \tiny{($\pm$ 0.3)} & {\bf 0.7} \tiny{($\pm$ 0.1)} & 7.2 \tiny{($\pm$ 0.5)} & 59.5 \tiny{($\pm$ 0.3)} & 30.4 \tiny{($\pm$ 0.5)} & 48.0 \tiny{($\pm$ 0.1)} & \underline{1.9} \tiny{($\pm$ 0.1)} & -0.4 \tiny{($\pm$ 0.2)} & 49.2 \tiny{($\pm$ 0.2)} & 30.2 \tiny{($\pm$ 0.2)} \\

POCON$^{\mathsection}$
& 50.6 \tiny{($\pm$ 0.7)} & 3.2 \tiny{($\pm$ 1.0)} & \underline{9.3} \tiny{($\pm$ 0.4)} & 59.3 \tiny{($\pm$ 0.3)} & 30.6 \tiny{($\pm$ 0.3)} & 45.2 \tiny{($\pm$ 0.4)} & 3.0 \tiny{($\pm$ 0.6)} & 2.7 \tiny{($\pm$ 0.4)} & 46.8 \tiny{($\pm$ 0.3)} & 28.8 \tiny{($\pm$ 0.1)} \\

\rowcolor{lightgray} Osiris-D (Ours) & {\bf 53.0} \tiny{($\pm$ 0.2)} & \underline{1.6} \tiny{($\pm$ 0.5)} & 8.4 \tiny{($\pm$ 0.3)} & {\bf 60.5} \tiny{($\pm$ 0.1)} & \underline{31.7} \tiny{($\pm$ 0.3)} & {\bf 50.1} \tiny{($\pm$ 0.2)} & 2.3 \tiny{($\pm$ 0.2)} & 4.2 \tiny{($\pm$ 0.3)}& {\bf 51.3} \tiny{($\pm$ 0.1)} & {\bf 31.3} \tiny{($\pm$ 0.4)} \\

\midrule

Offline & 52.5 \tiny{($\pm$ 0.4)} & - & - & 60.0 \tiny{($\pm$ 0.4)} & - & 52.5 \tiny{($\pm$ 0.4)} & - & - & 53.9 \tiny{($\pm$ 0.2)} & -\\

\bottomrule
\end{tabular}
}
\end{center}
\end{table}

\begin{table}[t]
\caption{Results on 10-task sequences on structured CIFAR-100 and Tiny-ImageNet. The two best models are marked. \texttt{Osiris-D} performs the best, surpassing \texttt{Offline} on Structured Tiny-ImageNet.
}
\label{tab:structured}
\begin{center}
\resizebox{1\linewidth}{!}{
\setlength{\tabcolsep}{4pt}
\begin{tabular}{l|ccccc|ccccc}
\toprule

& \multicolumn{5}{c|}{\textsc{Structured CIFAR-100}} & \multicolumn{5}{c}{\textsc{Structured Tiny-ImageNet}} \\

& \bf A ($\uparrow$) & \bf F ($\downarrow$) & \bf K ($\uparrow$) & \bf C ($\uparrow$) & \bf T ($\uparrow$) & \bf A ($\uparrow$) & \bf F ($\downarrow$) & \bf K ($\uparrow$) & \bf C ($\uparrow$) & \bf T ($\uparrow$) \\

\midrule

FT & 45.0 \tiny{($\pm$ 0.6)} & 5.5 \tiny{($\pm$ 0.4)} & 7.8 \tiny{($\pm$ 0.3)} & 59.2 \tiny{($\pm$ 0.5)} & 28.7 \tiny{($\pm$ 0.2)} & 34.2 \tiny{($\pm$ 0.2)} & 4.7 \tiny{($\pm$ 0.1)} & 5.7 \tiny{($\pm$ 0.4)} & 43.5 \tiny{($\pm$ 0.2)} & 26.0 \tiny{($\pm$ 0.2)} \\

\midrule

LUMP
& 48.5 \tiny{($\pm$ 0.7)} & \underline{3.6} \tiny{($\pm$ 0.6)} & 7.9 \tiny{($\pm$ 0.5)} & \underline{62.7} \tiny{($\pm$ 0.7)} & 29.7 \tiny{($\pm$ 0.6)} & 36.0 \tiny{($\pm$ 1.0)} & \underline{2.9} \tiny{($\pm$ 1.1)} & 5.3 \tiny{($\pm$ 0.2)} & \underline{45.2} \tiny{($\pm$ 0.8)} & 26.2 \tiny{($\pm$ 0.4)} \\

CaSSLe
& 47.4 \tiny{($\pm$ 0.3)} & {\bf 1.9} \tiny{($\pm$ 0.3)} & 4.1 \tiny{($\pm$ 0.2)} & 61.6 \tiny{($\pm$ 0.1)} & 30.0 \tiny{($\pm$ 0.3)} & 35.4 \tiny{($\pm$ 0.3)} & {\bf 2.1} \tiny{($\pm$ 0.2)} & 3.2 \tiny{($\pm$ 0.1)} & 43.9 \tiny{($\pm$ 0.3)} & 26.7 \tiny{($\pm$ 0.1)} \\

POCON$^{\mathsection}$
& 45.8 \tiny{($\pm$ 0.5)} & 5.2 \tiny{($\pm$ 0.3)} & 7.7 \tiny{($\pm$ 0.3)} & 59.6 \tiny{($\pm$ 0.8)} & 29.3 \tiny{($\pm$ 0.4)} & 34.4 \tiny{($\pm$ 0.6)} & 3.8 \tiny{($\pm$ 0.8)} & {\bf 5.5} \tiny{($\pm$ 0.2)} & 43.6 \tiny{($\pm$ 0.5)} & 25.7 \tiny{($\pm$ 0.1)} \\

\rowcolor{lightgray} Osiris-R (Ours) & \underline{49.0} \tiny{($\pm$ 0.4)} & 5.0 \tiny{($\pm$ 0.6)} & {\bf 8.3} \tiny{($\pm$ 0.6)} & \underline{62.8} \tiny{($\pm$ 0.2)} & {\bf 31.7} \tiny{($\pm$ 0.1)} & \underline{36.3} \tiny{($\pm$ 0.1)} & 3.6 \tiny{($\pm$ 0.2)} & {\bf 5.5} \tiny{($\pm$ 0.1)} & \underline{45.1} \tiny{($\pm$ 0.1)} & \underline{27.5} \tiny{($\pm$ 0.3)} \\

\rowcolor{lightgray} Osiris-D (Ours) & {\bf 49.8} \tiny{($\pm$ 0.1)} & 4.4 \tiny{($\pm$ 0.3)} & {\bf 8.4} \tiny{($\pm$ 0.2)} & {\bf 64.2} \tiny{($\pm$ 0.1)} & {\bf 31.5} \tiny{($\pm$ 0.2)} & {\bf 37.5} \tiny{($\pm$ 0.4)} & \underline{2.9} \tiny{($\pm$ 0.2)} & 5.0 \tiny{($\pm$ 0.4)} & {\bf 46.5} \tiny{($\pm$ 0.2)} & {\bf 28.1} \tiny{($\pm$ 0.1)} \\

\midrule

Offline & 52.5 \tiny{($\pm$ 0.4)} & - & - & 67.9 \tiny{($\pm$ 0.1)} & - & 36.8 \tiny{($\pm$ 0.1)} & - & - & 46.4 \tiny{($\pm$ 0.2)} & - \\

\bottomrule
\end{tabular}
}
\end{center}
\end{table}

\subsection{Results on the Standard Benchmarks} 
\label{sec:exp:standard}

In Table~\ref{tab:random}, we report the results of all models on 
Split-CIFAR-100.
Our first observation is that \texttt{FT} is already a very strong baseline with a final accuracy only 2\% below the \texttt{Offline} model's accuracy on the 5-task sequence. 
Moreover, \texttt{Osiris-D} closes this gap with \texttt{Offline} on both overall accuracy ($p=0.185$ by an unpaired t-test) and consolidation score. 
We hypothesize that 
this is in part attributed to that \texttt{Osiris-D} leverages the 
data ordering 
by contrasting the current task and the memory, although task labels are not explicitly given. On the other hand, \texttt{Offline} does not have any task label information.
Among all methods, \texttt{Osiris} consistently achieves the highest overall accuracy, consolidation score, and forward transfer, regardless of the number of tasks.
Comparing \texttt{Osiris-R} and \texttt{Osiris-D}, we find that there's still a trade-off between plasticity and stability.
\texttt{Osiris-R} shows more knowledge gain at the expense of higher forgetting, and \texttt{Osiris-D} shows lower forgetting but sometimes lower knowledge gain.

\texttt{CaSSLe} shows low forgetting on both task sequences but lower knowledge gains than all the other methods except \texttt{EWC}, indicating that parallel projectors (in \texttt{Osiris}) might be a better choice than sequential ones (in \texttt{CaSSLe}) at improving plasticity.
\texttt{POCON} is designed to maximize plasticity by distilling the model with a single-task expert, achieving the highest knowledge gain among the distillation-based methods (except \texttt{Osiris-D} on the 20 tasks). 
\texttt{LUMP} improves over \texttt{FT} on the 20-task but not on the 5-task sequence, and we hypothesize that it is because memory is more important when individual tasks are less diverse. 
Interestingly, the knowledge gain of \texttt{FT} is not the upper bound for UCL methods. This may be because UCL methods implicitly leverage learned representations or memory to help learn a new task (a form of forward transfer). 
For example, the methods with the highest knowledge gain on each benchmark are 
\texttt{ER+} 
and \texttt{ER++}, which both involve contrasting the current task and the memory, a component that is not present in \texttt{ER}. 

\paragraph{General findings.}
Similar to~\cite{madaan2022representational}, we find all UCL methods exhibit very low forgetting compared to previously reported numbers in SCL~\citep{chaudhry2018riemannian}. 
In general, the models with the lowest forgetting applies distillation.
Indeed, a previously hypothesized criticism for replay-based methods is that they are prune to overfitting to memory~\citep{fini2022self}, which we analyze empirically in Sec.~\ref{sec:exp:analysis}.
On the other hand, they show more plasticity and have higher knowledge gain on the new task.

\subsection{Results on the Structured Benchmarks}
\label{sec:exp:structured}

We show the results of recent UCL methods 
on the structured benchmarks in Table~\ref{tab:structured}.
On Structured CIFAR-100, the methods show higher forgetting than on both random 5-task and 20-task sequences (a caveat for such a comparison is that here we have ten tasks).
Nevertheless, \texttt{CaSSLe} and \texttt{LUMP} show relatively low forgetting but fail to address some of the other components of our framework~(Eq.~\ref{eq:framework}). 
All UCL methods improve over \texttt{FT},
and the two variants of \texttt{Osiris} outperform others in terms of knowledge gain, consolidation, forward transfer, and overall accuracy. This indicates that \texttt{Osiris} is robust to correlated task sequences.

On Structured Tiny-ImageNet, \texttt{FT} shows the highest knowledge gain, which means it benefits more from intra-task similarity. 
We hypothesize that contrastive learning benefits from high intra-task similarity because it provides hard negative pairs.
\texttt{Osiris} again outperforms other UCL methods in terms of knowledge gain, consolidation, and forward transfer.
Surprisingly, \texttt{Osiris-D} obtains better accuracy than \texttt{Offline} ($p=0.008$ by an unpaired t-test).
From a curriculum-learning perspective,
this 
suggests that the realistic, hierarchical structure offers a better task ordering than 
randomly constructed scenarios.
We explore how such a task ordering affects the representation structure in the next section.

\subsection{Analysis}
\label{sec:exp:analysis}

\paragraph{Balancing stability, plasticity, and consolidation.}

\begin{figure}[t]
\begin{center}
\begin{subfigure}[t]{0.74\textwidth}
    \includegraphics[width=\linewidth]{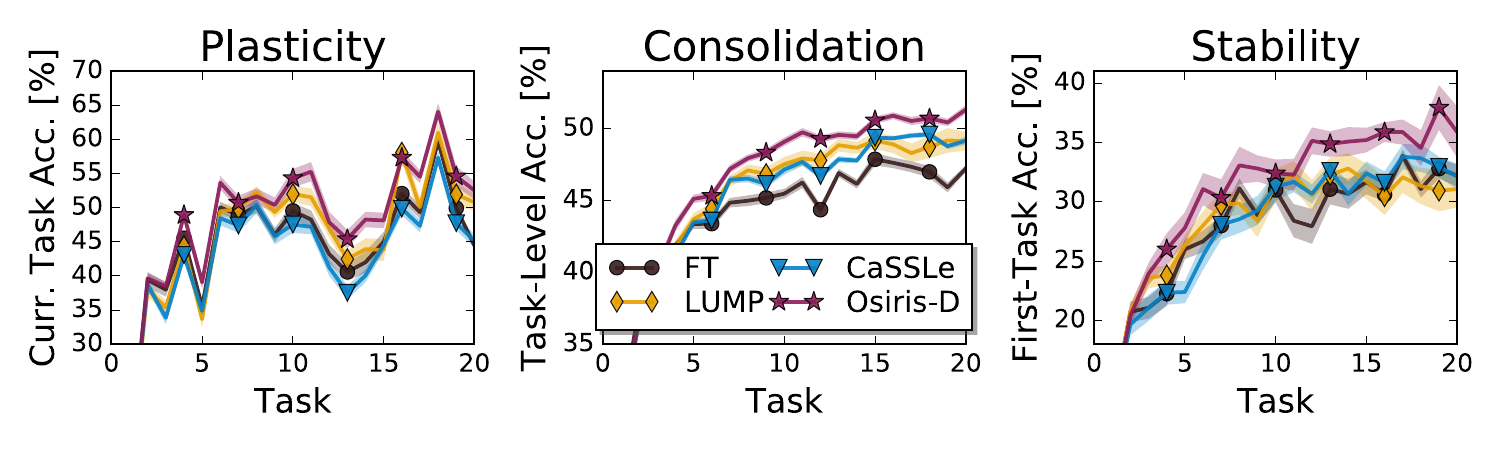}
    \caption{}
    \label{fig:exp:balance}
\end{subfigure} %
\begin{subfigure}[t]{0.25\textwidth}
    \includegraphics[width=\linewidth]{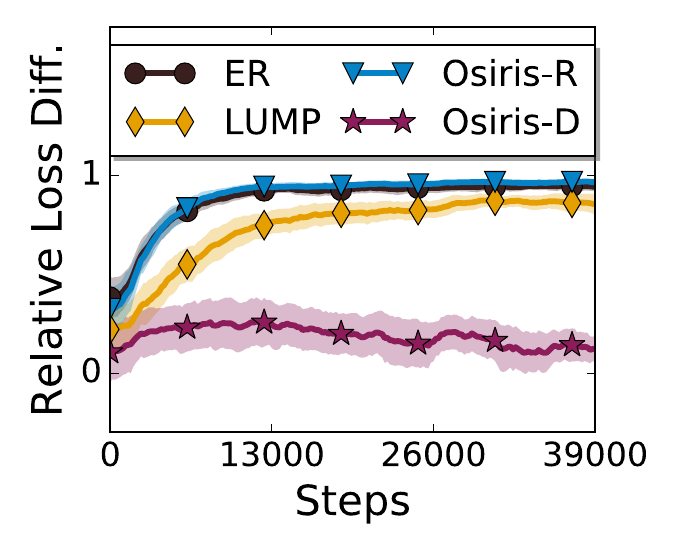}
    \caption{}
    \label{fig:exp:overfit}
\end{subfigure} %
\end{center}
\caption{(a) Interplay between plasticity (current-task accuracy), cross-task consolidation (task-level KNN accuracy), and stability (accuracy of the first task throughout training). \texttt{Osiris-D} balances the three aspects well and is usually the top performer.
(b) 
Relative 
difference between the contrastive loss on past-task data and on memory for replay-based methods. 
All methods except \texttt{Osiris-D} show signs of overfitting.
}
\label{fig:exp:tradeoff}
\end{figure}

We now examine how UCL methods balance plasticity, cross-task consolidation, and stability. 
We use \texttt{Osiris-D} for our analysis in this section since \texttt{Osiris-R} shows similar behavior. Similar to the analysis provided by~\cite{gomez2024plasticity}, in Fig.~\ref{fig:exp:tradeoff},
we plot for different UCL methods the current task accuracy (plasticity), task-level KNN accuracy (cross-task consolidation), and accuracy of the first task (stability) throughout training on standard, 20-task Split-CIFAR-100. We plot the accuracy for all the other tasks in Appendix~\ref{sec:appendix:plots}. 
The first observation is that \texttt{Osiris} performs relatively well on all three aspects throughout training. 
\texttt{LUMP} and \texttt{CaSSLe} have similar overall accuracy in Table~\ref{tab:random}. They show the same level of cross-task consolidation in Fig.~\ref{fig:exp:tradeoff} because they do not directly enforce it. Among the two, \texttt{LUMP} shows higher plasticity but lower stability near the end of training. 
Both methods show better overall accuracy than \texttt{FT}, which may be attributed to their better consolidation scores.

\begin{table}[t]
\caption{Ablation of Osiris-D's components on 20-task Split-CIFAR-100. Isolating feature spaces is crucial for high knowledge gain; $\mathcal{L}_\text{cross}$ is important for knowledge gain and separation of task representations; $\mathcal{L}_\text{past}$ helps reduce forgetting.}
\label{tab:ablation}
    \centering
    \resizebox{0.75\linewidth}{!}{
    \setlength{\tabcolsep}{4pt}
    \begin{tabular}{l|ccccc}
    \toprule
    &\bf A ($\uparrow$) & \bf F ($\downarrow$) & \bf K ($\uparrow$) & \bf C ($\uparrow$) & \bf T ($\uparrow$)  \\
    \midrule
    w/o isolated space & 47.8 \tiny{($\pm$ 0.3)} & 2.8 \tiny{($\pm$ 0.3)} & 2.0 \tiny{($\pm$ 0.4)} & 49.3 \tiny{($\pm$ 0.2)} & 30.4 \tiny{($\pm$ 0.3)} \\
    
    w/o $\mathcal{L}_\text{cross}$ & 45.9 \tiny{($\pm$ 0.1)} & 2.5 \tiny{($\pm$ 0.3)} & 1.8 \tiny{($\pm$ 0.2)} & 47.2 \tiny{($\pm$ 0.2)} & 29.2 \tiny{($\pm$ 0.2)} \\
    
    w/o $\mathcal{L}_\text{past}$ & 49.2 \tiny{($\pm$ 0.4)} & 2.6 \tiny{($\pm$ 0.3)} & {\bf 4.8} \tiny{($\pm$ 0.5)} & 50.4 \tiny{($\pm$ 0.3)} & 30.5 \tiny{($\pm$ 0.2)} \\
    
    Full & {\bf 50.1} \tiny{($\pm$ 0.2)} & {\bf 2.3} \tiny{($\pm$ 0.2)} & 4.2 \tiny{($\pm$ 0.3)}& {\bf 51.3} \tiny{($\pm$ 0.1)} & {\bf 31.3} \tiny{($\pm$ 0.4)} \\
    \bottomrule
    \end{tabular}}
    
\end{table}
\begin{table}[t]
\caption{Within-environment accuracy on Structured Tiny-ImageNet. Both \texttt{FT} and \texttt{Osiris-D} outperforms \texttt{Offline} on Environment 3, but only \texttt{Osiris-D} achieves performance similar to \texttt{Offline} on Env. 1 and 2.}
\label{tab:in_env_acc}
    \centering
    \resizebox{0.5\linewidth}{!}{
    {\renewcommand{\arraystretch}{1.12}
    \begin{tabular}{l|ccc}
    \toprule
    Method & Env. 1 & Env. 2 & Env. 3 \\
    \midrule
    Offline & 44.2 \tiny{($\pm$ 0.6)} & 60.9 \tiny{($\pm$ 0.7)} & 50.3 \tiny{($\pm$ 0.5)} \\
    FT & 39.0 \tiny{($\pm$ 0.2)} & 57.1 \tiny{($\pm$ 0.4)} & 54.7 \tiny{($\pm$ 0.5)} \\
    Osiris-D & 43.7 \tiny{($\pm$ 0.6)} & 60.2 \tiny{($\pm$ 0.7)} & 55.1 \tiny{($\pm$ 0.5)} \\
    \bottomrule
    \end{tabular}}
    }
\end{table}

\begin{figure}[t]
\begin{center}
\includegraphics[width=1\linewidth]{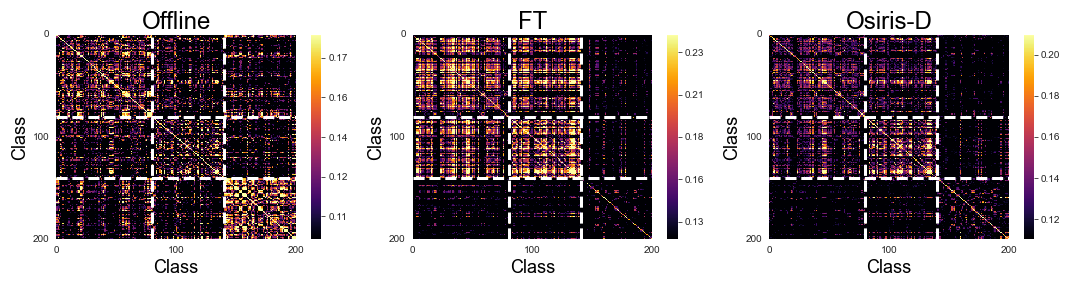}
\end{center}
\caption{
Mean cosine similarity between pairs of examples drawn from pairs of classes. Environment switches are marked with dashed white lines. Classes within the third environment are projected to nearby positions on the representation space by \texttt{Offline}, but not by \texttt{FT} and \texttt{Osiris-D}. 
}
\label{fig:exp:feature_similarity}
\end{figure}

\paragraph{Ablations.} In Table~\ref{tab:ablation}, we show the results of our framework after removing each component. When using a shared projector for all three losses, the model's knowledge gain drops from 4.2\% to 2.0\%, which shows that using separate spaces helps plasticity, as we have hypothesized. When not using $\mathcal{L}_{\text{cross}}$, the model shows the lowest cross-task discrimination score among all the models being compared here. It also shows low forgetting because the only loss applied on output space of $h \circ f$ is now the distillation loss ($\mathcal{L}_{\text{past}}$). On the other hand, after removing $\mathcal{L}_{\text{past}}$, the model exhibits 
the highest knowledge gain, which means that learning the new task gains benefit from $\mathcal{L}_{\text{cross}}$. Finally, with all the components, \texttt{Osiris} balances all three aspects of our framework and achieves the best scores on all metrics except knowledge gain,
without requiring manually adjusting the loss weights.

\paragraph{How does a structured task sequence affect representation?}

In Table~\ref{tab:in_env_acc}, we show the within-environment accuracy for \texttt{Offline}, \texttt{FT}, and \texttt{Osiris-D}. Both \texttt{FT} and \texttt{Osiris} perform better than \texttt{Offline} on the last environment but not as well on the first two. Compared to \texttt{FT}, \texttt{Osiris} shows less forgetting and performs better in 
previously-observed environments.
To examine their representation geometry, we plot the mean cosine similarity matrix between features of examples from pairs of classes, for these three methods, in Fig.~\ref{fig:exp:feature_similarity}.
The matrix shows that classes within the last environment are naturally projected by \texttt{Offline} to nearby positions on the representation hypersphere. Differently, \texttt{FT} and \texttt{Osiris} can better distinguish between classes in the third environment. Among the two, \texttt{Osiris} distinguishes between different classes within the first two environments better.
Together with within-environment accuracy, this provides evidence that UCL methods benefit from the ordered task sequence such that they distinguish examples in the last environment better. At the same time, \texttt{Offline} is less sensitive distinguishing these examples.

\paragraph{Does replay-based methods overfit to memory?}

It has been shown that replay-based methods can overfit to memory in SCL~\citep{verwimp2021rehearsal, buzzega2021rethinking}. The same phenomenon has been hypothesized by~\cite{fini2022self} to also exist in UCL, but not yet empirically verified. As SSL learns features that generalize better~\citep{madaan2022representational}, we test this hypothesis empirically.
In Fig.~\ref{fig:exp:overfit}, we plot for replay-based methods the 
relative difference between the contrastive loss of a batch sampled from all data observed ($\mathcal{L}_{\text{all}}$) and from the memory ($\mathcal{L}_{\text{mem}}$), both excluding any data from the current task. The relative difference is defined as $\frac{\mathcal{L}_{\text{all}}-\mathcal{L}_{\text{mem}}}{\mathcal{L}_{\text{all}}}$. 
A large difference could indicate the failure to generalize the representations learned from memory to past tasks, which defeats the purpose of replay.  
For \texttt{Osiris}, we plot the curves with $h \circ f$ here in Fig.~\ref{fig:exp:overfit} and for $g \circ f$ in Appendix~\ref{sec:appendix:plots}. In Fig.~\ref{fig:exp:overfit}, the curves for \texttt{ER}, \texttt{LUMP}, and \texttt{Osiris-R} increase at first and become relatively stable afterwards. Their final values are all much larger than zero,
indicating the possibility of overfitting to memory.
Since \texttt{Osiris-D} does not explicitly minimize the contrastive loss on the memory, it does not overfit to it and shows close-to-zero relative loss difference. This could explain why it always has lower forgetting than \texttt{Osiris-R}. 

While Fig.~\ref{fig:exp:overfit} could indicate that \texttt{Osiris-R} overfits to the memory with $h \circ f$, the effect of $\mathcal{L}_{\text{cross}}$ on the encoder $f$ appears to be less sensitive to the precision of the representations produced by $h \circ f$. In all of our results (Tables~\ref{tab:random},~\ref{tab:structured},~\ref{tab:wtl:random}, and~\ref{tab:wtl:structured}), \texttt{Osiris-R} and \texttt{Osiris-D} are consistently the best performers in consolidation scores, which leads to good overall accuracy. Additionally, neither \texttt{Osiris-R} nor \texttt{Osiris-D} overfits with the representations produced by $g \circ f$ as shown in Fig.~\ref{fig:appendix:overfit_curr}.
Overall, our findings empirically support the claim in prior work that using replay for UCL may cause the model to overfit to the memory~\citep{fini2022self}, but also show that we still need the memory to improve consolidation, which is crucial for performance.

\section{Related Work}
\label{sec:related}

\paragraph{Self-supervised learning.} A large body of work in SSL belongs to the contrastive learning family~\citep{chen2020simple, he2020momentum, Misra_2020_CVPR, hadsell2006dimensionality, tian2020contrastive, oord2018representation, wu2018unsupervised, gutmann2010noise}, which we focus on in this study. The main idea is to match the representations of two augmented views of the same image and repel the representations of different images to learn semantically meaningful representations. Clustering-based methods share high-level intuition but perform contrastive learning on the cluster level rather than the instance level \citep{caron2020unsupervised, he2016deep}. Other works have explored relaxing the need for negative pairs, usually by asymmetric architectures~\citep{grill2020bootstrap, chen2021exploring} or losses that enforce variance in representations~\citep{zbontar2021barlow, bardes2022vicreg, ermolov2021whitening}. SSL methods that work on transformers have also been proposed in recent years~\citep{he2022masked, caron2021emerging}.

\paragraph{Continual learning.} SCL methods are commonly partitioned into three categories. \emph{Regularization}-based methods~\citep{kirkpatrick2017overcoming, schwarz2018progress, chaudhry2018riemannian, zenke2017continual, aljundi2018memory, castro2018end, douillard2020podnet, hou2019learning, wu2019large} regularize model parameters such that they do not drift too far from previous optima. \emph{Replay}-based methods~\citep{buzzega2020dark, robins1995catastrophic, hayes2020remind, rebuffi2017icarl, chaudhry2018efficient, lopez2017gradient, NEURIPS2019_15825aee, ostapenko2019learning} use a memory buffer storing data from past task and use them for replay. Finally, \emph{architecture}-based methods~\citep{ostapenko2021continual, rusu2016progressive, serra2018overcoming, li2019learn} dynamically introduce new parameters for each task to reduce forgetting. Limited progress has been made in UCL. The area is first explored by~\cite{rao2019continual, smith2019unsupervised}, but their work is limited to small datasets such as handwritten digits and is hard to scale. Recent work focuses on improving SSL-based UCL~\citep{madaan2022representational, fini2022self, gomez2024plasticity, gomez2022continually, cheng2023contrastive}. In contrast, we investigate what features are essential in UCL and offer new insights and practical guidance around normalization. SSL also helps SCL~\citep{cha2021co2l, caccia2021new} by improving the model's representations. It is worth mentioning that properties of representations learned with SSL in continual learning have been empirically studied~\citep {davari2022probing, galashov2023continually, gallardo2021self}.

\section{Conclusion}

This work identifies three key components in UCL for integrating representation learning in the present and the past tasks: plasticity, stability and cross-task consolidation. Existing methods fall under our unifying framework by optimizing only a subset of objectives, whereas our proposed method \texttt{Osiris} explicitly optimizes and balances all three desiderata. \texttt{Osiris} achieves state-of-the-art performance on all UCL benchmarks and shows 
better accuracy on our realistic Structured Tiny-ImageNet benchmark than offline iid training. Our work sheds new light on the potential learning mechanisms of continual learning agents in the real world. 
Future work 
will
extend our framework to non-contrastive SSL approaches and evaluate more realistic learning environments, such as lifelong video recordings.

\section*{Acknowledgment}
We appreciate the constructive feedback from five anonymous reviewers. We also thank Lucas Caccia, Siddarth Venkatraman, and Michael Chong Wang for helpful discussions and Avery Ryoo for pointing out some typos in earlier drafts. LC and YZ acknowledge the generous support of the CIFAR AI Chair program. This work obtained support by the funds provided by the National Science Foundation and by DoD OUSD (R\&E) under Cooperative Agreement PHY-2229929 (The NSF AI Institute for Artificial and Natural Intelligence). This research was also enabled in part by compute resources provided by Mila (mila.quebec).

{
\setstretch{0.98}
\bibliography{ref}
\bibliographystyle{apalike}
}

\appendix
\newpage

\section{Proof of Proposition 1}
\label{sec:appendix:lump_linearity}

\begin{proof}
Let $\nu \sim \text{Beta}(\alpha, \alpha)$.
Define $\vz_i \coloneqq g(f(\vx_i)) / \|g(f(\vx_i))\|_2$, $\vu_i \coloneqq g(f(\vy_i)) / \|g(f(\vy_i))\|_2$, and $\Tilde{\vz_i} \coloneqq g(f(\Tilde{\vx_i})) / \|g(f(\Tilde{\vx_i}))\|_2$ for all $i \in \{0, \ldots |X|\}$. Assume that the representations are also linearly mixed in the representation space, 
i.e., $\Tilde{\vz_i} = \nu \vz_i + (1-\nu) \vu_i$.

Let $\Tilde{\vz_i}$ be the anchor, the representation of its other augmented view $\Tilde{\vz_i}'$ as the positive, and a set of representations of mixed examples $\{\Tilde{\vz_j}\}_{j\neq i}$ as negatives.
Then we can express the \texttt{LUMP} loss as
\begin{equation}
\label{eq:lump_loss}
\mathcal{L}^{\text{LUMP}}(X, Y; \nu, f_\Theta, g_\Phi) 
=
\mathbb{E}_\nu \left[ 
-\log \frac{\exp(\Tilde{\vz_i}^\top \Tilde{\vz_i}')}{\exp(\Tilde{\vz_i}^\top \Tilde{\vz_i}') + \sum_{j\neq i} \exp(\Tilde{\vz_i}^\top \Tilde{\vz_j})} \right] \, .
\end{equation}

For convenience, define scalar quantities $p_i$, $p_j$'s as the softmax probability of predictions:
\begin{align}
p_{i} 
&= 
\frac{\exp(\Tilde{\vz_i}^\top \Tilde{\vz_i}')}{\exp(\Tilde{\vz_i}^\top \Tilde{\vz_i}') + \sum_{j\neq i} \exp(\Tilde{\vz_i}^\top \Tilde{\vz_j})}
\, , \quad \text{and} \label{eq:softmax_i}\\
p_{j} 
&= 
\frac{\exp(\Tilde{\vz_i}^\top \Tilde{\vz_j})}{\exp(\Tilde{\vz_i}^\top \Tilde{\vz_i}') + \sum_{k\neq i} \exp(\Tilde{\vz_i}^\top \Tilde{\vz_k})}
\, , \quad \text{if} \; j \neq i \, , \label{eq:softmax_j}
\end{align}
where $0 \leq p_i, p_j \leq 1$. Then gradient of the \texttt{LUMP} loss in Eq.~\ref{eq:lump_loss} w.r.t. the anchor is
\begin{align}
\mathbb{E}_\nu \left[ \frac{\partial \mathcal{L}^{\text{LUMP}}}{\partial \Tilde{\vz_i}} \right] 
&=
\mathbb{E}_\nu \left[ (p_{i} - 1) \Tilde{\vz_i}'
+
\sum_{j\neq i} p_{j} \Tilde{\vz_j}
\right]  \label{eq:mixed} \\
&=
\mathbb{E}_\nu \left[ (p_{i} - 1) 
\left(
\nu \vz_i' + (1-\nu) \vu_i'
\right)
+
\sum_{j\neq i} p_{j} 
\left(
\nu \vz_j + (1-\nu) \vu_j
\right)
\right] \label{eq:expanded} \\
&=
\mathbb{E}_\nu \left[ 
(p_{i}-1)(\nu) \vz_i'
+
(p_{i}-1)(1-\nu) \vu_i'
+
\sum_{j\neq i}
(p_{j})(\nu) \vz_j
+
\sum_{j\neq i} (p_{j})(1-\nu) \vu_j
\right] 
\, ,
\end{align}
where the equality holds between lines~\ref{eq:mixed} and~\ref{eq:expanded} 
because of our linearity assumption.
Therefore, the gradient w.r.t. the current-task example's representation is
\begin{align}
&\mathbb{E}_\nu \left[ \frac{\partial \mathcal{L}^{\text{LUMP}}}{\partial \vz_i} \right] 
=
\mathbb{E}_\nu \left[ \frac{\partial \mathcal{L}^{\text{LUMP}}}{\partial \Tilde{\vz_i}}\frac{\partial \Tilde{\vz_i}}{\partial \vz_i} 
\right] 
=
\mathbb{E}_\nu \left[ \nu \frac{\partial \mathcal{L}^{\text{LUMP}}}{\partial \Tilde{\vz_i}}
\right] \\
&=
\mathbb{E}_\nu \left[ 
(p_{i}-1)(\nu^2) \vz_i'
+
(p_{i}-1)(1-\nu)(\nu) \vu_i'
+
\sum_{j\neq i}
(p_{j})(\nu^2) \vz_j
+
\sum_{j\neq i} (p_{j})(1-\nu)(\nu) \vu_j
\right] \\
&=
(p_{i}-1)\mathbb{E}[\nu^2] \vz_i'
+
(p_{i}-1)\left(\mathbb{E}[\nu] - \mathbb{E}[\nu^2]\right) \vu_i'
+
\sum_{j\neq i}
(p_{j})\mathbb{E}[\nu^2] \vz_j
+
\sum_{j\neq i} (p_{j})\left(\mathbb{E}[\nu] - \mathbb{E}[\nu^2]\right) \vu_j \\
&=
(p_{i}-1)\mathbb{E}[\nu^2] \vz_i'
+
\sum_{j\neq i}
(p_{j})\mathbb{E}[\nu^2] \vz_j
+
\sum_{j\neq i} (p_{j})\left(\mathbb{E}[\nu] - \mathbb{E}[\nu^2]\right) \vu_j
+
(p_{i}-1)\left(\mathbb{E}[\nu] - \mathbb{E}[\nu^2]\right) \vu_i'
\, .
\end{align}

Notice that this is a weighted sum of $\vz_i'$, $\{\vz_j\}_j$, $\{\vu_j\}_j$, and $\vu_i'$. The first and second terms are the weighted representations of views of $\vx \in X$ the other two terms are the weighted representations of views of $\vy \in Y$. 

Now, the coefficients involve $p_{i}, p_{j}$, which are the softmax probabilities given by Eq.~\ref{eq:softmax_i} and Eq.~\ref{eq:softmax_j}; they also involve $\mathbb{E}[\nu], \mathbb{E}[\nu^2]$, which are the first and second moments of Beta($\alpha$, $\alpha$) where $\mathbb{E}[\nu^2], \mathbb{E}[\nu] - \mathbb{E}[\nu^2] > 0$. Therefore, it is easy to see that the coefficients on $\vz_i', \vu_i'$ are negative, and the coefficients on $\{\vz_j\}_j$ and $\{\vu_j\}_j$ are positive. 
This conclusion holds as long as $\mathbb{E}[\nu^2], \mathbb{E}[\nu] - \mathbb{E}[\nu^2] > 0$ are satisfied; i.e., $\nu$ does not necessarily need to be sampled from a symmetric Beta distribution as chosen by \texttt{LUMP}.

This concludes our proof for the first equality. The proof for $\mathbb{E}_\nu \left[ \frac{\partial \mathcal{L}^{\text{LUMP}}}{\partial \vu_i} \right]$
is similar due to symmetry.
\end{proof}

\section{Implementation Details}
\label{sec:appendix:implementation}

\paragraph{General details.} For GroupNorm, we set the number of groups to $\text{min}(32, \floor{\#\text{channels}/4})$. We replace the ReLU activation functions following GroupNorm with Mish activation~\citep{misra2019mish} to avoid dying ReLUs, or \emph{elimination singularities}~\citep{qiao2019micro}. We use the same data augmentation procedure and parameters as~\cite{zbontar2021barlow, grill2020bootstrap},
and change the resized image size to $32\times32$ for CIFAR-100 and $64\times64$ for Tiny-ImageNet. We set $\tau=0.1$ for all contrastive losses. We train all our models for $200$ epochs with SGD optimizer, learning rate of 0.03, weight decay of $5e^{-4}$, and batch size of $|X|=256$ on all benchmarks, following~\cite{madaan2022representational}. We do not use LARS~\citep{you2017large} because we do not find it make significant differences on batches of size 256. We train \texttt{Offline} for the same number of steps as continual models. The models are trained on two NVIDIA Quadro RTX 8000 GPUs for all experiments. For KNN evaluation, we follow \cite{he2020momentum, wu2018unsupervised}'s set up, where we set $k=200$ and temperature $\tau_{\text{KNN}}=0.1$. 

\paragraph{UCL methods.} 
All hyperparameters are fine-tuned manually with at least three values. 
We adapt the official implementations of \texttt{CaSSLe}, \texttt{LUMP} to our codebase. Since \texttt{POCON} is a concurrent work and its code has not been made public yet, we re-implement it. 
For online \texttt{EWC}, we normalize the diagonal Fisher information matrix for each task following the authors~\citep{schwarz2018progress} and set $\gamma=1$; we use $50$, $100$ as the weight on the regularization loss on the five and 20 task sequences of Split-CIFAR-100, respectively. For \texttt{LUMP}, we set the Beta distribution parameter $\alpha=0.4$ on all experiments; it's the same value used by the authors on Tiny-ImageNet, and we find it performs better than the authors' parameter $\alpha=0.1$ on CIFAR-100.
For online \texttt{POCON}, we save the model checkpoint every $ds=2,000$ steps for CIFAR-100 and $ds=4,000$ steps for Tiny-ImageNet; we set the weights to $1$ on all its loss terms.
For \texttt{DER}, \texttt{ER}, and \texttt{ER+}, we set the weights to $1$ on the additional loss terms. For all replay-based methods except for \texttt{LUMP} (which requires $|Y| = |X| = 256$), we uniformly sample $|Y|=\frac{3}{4}|X|=192$ examples from the memory at each step for replay.

\section{Evaluation Results Given Task Identities}
\label{sec:appendix:wtl}

\begin{table}[t]
\caption{Results on standard Split-CIFAR-100 with five or 20 tasks. The ground truth task identity for each example is given to the model at test time, so the model predicts the most probable class within the task. The best model in each column is made \textbf{bold} and the second-best model is \underline{underlined}. $^\dagger$Improved \texttt{DER}. $^{\mathsection}$Online version of \texttt{POCON}. We separate replay-based (top) and distillation-based methods (bottom) for easier comparisons.}
\label{tab:wtl:random}
\begin{center}
\resizebox{0.99\linewidth}{!}{
\setlength{\tabcolsep}{4pt}
\begin{tabular}{l|ccccc|ccccc}
\toprule
& \multicolumn{5}{c|}{\textsc{5-Task Split-CIFAR-100}} & \multicolumn{5}{c}{\textsc{20-Task Split-CIFAR-100}}\\
& 
\bf A ($\uparrow$) & \bf F ($\downarrow$) & \bf K ($\uparrow$) & \bf C ($\uparrow$) & \bf T ($\uparrow$) & 
\bf A ($\uparrow$) & \bf F ($\downarrow$) & \bf K ($\uparrow$) & \bf C ($\uparrow$) & \bf T ($\uparrow$) \\

\midrule

FT & 72.4 \tiny{($\pm$ 0.3)} & 1.7 \tiny{($\pm$ 0.3)} & 7.9 \tiny{($\pm$ 0.0)} & 59.1 \tiny{($\pm$ 0.2)} & 36.3 \tiny{($\pm$ 0.5)} & 84.4 \tiny{($\pm$ 0.1)} & 4.2 \tiny{($\pm$ 0.1)} & 6.6 \tiny{($\pm$ 0.0)} & 47.2 \tiny{($\pm$ 0.1)} & 28.6 \tiny{($\pm$ 0.1)} \\

\midrule

ER
& 72.9 \tiny{($\pm$ 0.2)} & 1.1 \tiny{($\pm$ 0.2)} & 6.9 \tiny{($\pm$ 0.3)} & 59.7 \tiny{($\pm$ 0.4)} & 36.9 \tiny{($\pm$ 0.3)} & 85.1 \tiny{($\pm$ 0.2)} & 3.0 \tiny{($\pm$ 0.2)} & 5.1 \tiny{($\pm$ 0.4)} & 48.2 \tiny{($\pm$ 0.5)} & 29.6 \tiny{($\pm$ 0.3)} \\

DER$^\dagger$
& 72.5 \tiny{($\pm$ 0.3)} & 1.6 \tiny{($\pm$ 0.4)} & {\bf 8.1} \tiny{($\pm$ 0.2)} & 59.0 \tiny{($\pm$ 0.4)} & 36.2 \tiny{($\pm$ 0.4)} & 84.2 \tiny{($\pm$ 0.2)} & 4.2 \tiny{($\pm$ 0.2)} & \underline{6.7} \tiny{($\pm$ 0.5)} & 47.2 \tiny{($\pm$ 0.2)} & 28.5 \tiny{($\pm$ 0.2)} \\

LUMP
& 71.2 \tiny{($\pm$ 0.5)} & 0.9 \tiny{($\pm$ 0.5)} & 5.8 \tiny{($\pm$ 0.4)} & 58.4 \tiny{($\pm$ 0.4)} & 35.8 \tiny{($\pm$ 0.3)} & 85.3 \tiny{($\pm$ 0.6)} & 2.4 \tiny{($\pm$ 0.7)} & 4.3 \tiny{($\pm$ 0.3)} & 49.1 \tiny{($\pm$ 1.0)} & 29.3 \tiny{($\pm$ 0.1)} \\

ER+ & 73.0 \tiny{($\pm$ 0.5)} & 1.7 \tiny{($\pm$ 0.3)} & 7.9 \tiny{($\pm$ 0.2)} & \underline{60.1} \tiny{($\pm$ 0.3)} & 36.8 \tiny{($\pm$ 0.5)} & 85.1 \tiny{($\pm$ 0.2)} & 3.9 \tiny{($\pm$ 0.1)} & 6.4 \tiny{($\pm$ 0.2)} & 48.0 \tiny{($\pm$ 0.1)} & 29.2 \tiny{($\pm$ 0.1)} \\

ER++ & 72.5 \tiny{($\pm$ 0.5)} & 1.3 \tiny{($\pm$ 0.6)} & 7.0 \tiny{($\pm$ 0.0)} & 59.8 \tiny{($\pm$ 0.3)} & 36.8 \tiny{($\pm$ 0.4)} & 85.3 \tiny{($\pm$ 0.2)} & 2.8 \tiny{($\pm$ 0.2)} & 4.9 \tiny{($\pm$ 0.1)} & 49.0 \tiny{($\pm$ 0.5)} & 29.6 \tiny{($\pm$ 0.0)} \\

\rowcolor{lightgray} Osiris-R (Ours) & \underline{73.3} \tiny{($\pm$ 0.2)} & 1.3 \tiny{($\pm$ 0.3)} & 6.7 \tiny{($\pm$ 0.4)} & \underline{60.1} \tiny{($\pm$ 0.1)} & {\bf 37.7} \tiny{($\pm$ 0.1)} & \underline{86.5} \tiny{($\pm$ 0.2)} & 2.5 \tiny{($\pm$ 0.2)} & 5.4 \tiny{($\pm$ 0.5)} & \underline{50.5} \tiny{($\pm$ 0.5)} & {\bf 30.2} \tiny{($\pm$ 0.4)} \\

\midrule

EWC
& 65.6 \tiny{($\pm$ 0.7)} & 1.4 \tiny{($\pm$ 0.8)} & 3.7 \tiny{($\pm$ 0.5)} & 53.9 \tiny{($\pm$ 0.7)} & 32.8 \tiny{($\pm$ 0.7)} & 79.0 \tiny{($\pm$ 0.2)} & {\bf 1.9} \tiny{($\pm$ 0.1)} & 3.7 \tiny{($\pm$ 0.1)} & 39.4 \tiny{($\pm$ 0.3)} & 24.0 \tiny{($\pm$ 0.2)} \\

CaSSLe
& 72.6 \tiny{($\pm$ 0.2)} & {\bf 0.5} \tiny{($\pm$ 0.4)} & 6.7 \tiny{($\pm$ 0.5)} & 59.5 \tiny{($\pm$ 0.3)} & 36.1 \tiny{($\pm$ 0.4)} & 85.4 \tiny{($\pm$ 0.1)} & 2.5 \tiny{($\pm$ 0.1)} & 5.1 \tiny{($\pm$ 0.2)} & 49.2 \tiny{($\pm$ 0.2)} & 28.8 \tiny{($\pm$ 0.2)} \\

POCON$^{\mathsection}$
& 72.1 \tiny{($\pm$ 0.7)} & 2.2 \tiny{($\pm$ 0.8)} & {\bf 8.0} \tiny{($\pm$ 0.5)} & 59.3 \tiny{($\pm$ 0.3)} & 36.4 \tiny{($\pm$ 0.4)} & 84.0 \tiny{($\pm$ 0.2)} & 4.4 \tiny{($\pm$ 0.4)} & {\bf 6.9} \tiny{($\pm$ 0.3)} & 46.8 \tiny{($\pm$ 0.3)} & 28.3 \tiny{($\pm$ 0.3)} \\

\rowcolor{lightgray} Osiris-D (Ours) & {\bf 73.9} \tiny{($\pm$ 0.1)} & \underline{0.7} \tiny{($\pm$ 0.3)} & 6.7 \tiny{($\pm$ 0.5)} & {\bf 60.5} \tiny{($\pm$ 0.1)} & \underline{37.3} \tiny{($\pm$ 0.2)} & {\bf 86.7} \tiny{($\pm$ 0.3)} & \underline{2.2} \tiny{($\pm$ 0.3)} & 5.1 \tiny{($\pm$ 0.2)} & {\bf 51.3} \tiny{($\pm$ 0.1)} & \underline{30.0} \tiny{($\pm$ 0.2)} \\

\midrule

Offline & 74.0 \tiny{($\pm$ 0.2)} & - & - & 60.4 \tiny{($\pm$ 0.4)} & - & 88.7 \tiny{($\pm$ 0.1)} & - & - & 53.9 \tiny{($\pm$ 0.2)} & -\\
\bottomrule
\end{tabular}
}
\end{center}
\end{table}

\begin{table}[t]
\caption{Results on the 10-task sequences on structured CIFAR-100 and Tiny-ImageNet. The ground truth task identity for each example is given to the model at test time, so the model predicts the most probable class within the task. The two best models are marked. \texttt{Osiris-D} performs the best, surpassing \texttt{Offline} on Structured Tiny-ImageNet.
}
\label{tab:wtl:structured}
\begin{center}
\resizebox{0.99\linewidth}{!}{
\setlength{\tabcolsep}{4pt}
\begin{tabular}{l|ccccc|ccccc}
\toprule
& \multicolumn{5}{c|}{\textsc{Structured CIFAR-100}} & \multicolumn{5}{c}{\textsc{Structured Tiny-ImageNet}} \\
& \bf A ($\uparrow$) & \bf F ($\downarrow$) & \bf K ($\uparrow$) & \bf C ($\uparrow$) & \bf T ($\uparrow$) & \bf A ($\uparrow$) & \bf F ($\downarrow$) & \bf K ($\uparrow$) & \bf C ($\uparrow$) & \bf T ($\uparrow$) \\

\midrule

FT & 64.6 \tiny{($\pm$ 0.8)} & 7.6 \tiny{($\pm$ 0.8)} & 10.0 \tiny{($\pm$ 0.2)} & 59.2 \tiny{($\pm$ 0.5)} & 29.9 \tiny{($\pm$ 0.3)} & 57.8 \tiny{($\pm$ 0.2)} & 6.2 \tiny{($\pm$ 0.3)} & 6.4 \tiny{($\pm$ 0.2)} & 43.5 \tiny{($\pm$ 0.2)} & 34.4 \tiny{($\pm$ 0.1)} \\

\midrule

LUMP
& 66.6 \tiny{($\pm$ 0.4)} & {\bf 3.7} \tiny{($\pm$ 0.2)} & 7.2 \tiny{($\pm$ 0.4)} & \underline{62.7} \tiny{($\pm$ 0.7)} & 30.6 \tiny{($\pm$ 0.1)} & 59.5 \tiny{($\pm$ 0.8)} & {\bf 2.9} \tiny{($\pm$ 1.1)} & 4.9 \tiny{($\pm$ 0.2)} & \underline{45.2} \tiny{($\pm$ 0.8)} & 34.1 \tiny{($\pm$ 0.3)} \\

CaSSLe
& 66.4 \tiny{($\pm$ 0.2)} & \underline{4.2} \tiny{($\pm$ 0.4)} & 7.7 \tiny{($\pm$ 0.3)} & 61.6 \tiny{($\pm$ 0.1)} & 30.5 \tiny{($\pm$ 0.1)} & 59.0 \tiny{($\pm$ 0.6)} & 4.3 \tiny{($\pm$ 0.3)} & 5.4 \tiny{($\pm$ 0.6)} & 43.9 \tiny{($\pm$ 0.3)} & 34.7 \tiny{($\pm$ 0.3)} \\

POCON$^{\mathsection}$
& 64.5 \tiny{($\pm$ 0.4)} & 7.6 \tiny{($\pm$ 0.6)} & {\bf 9.8} \tiny{($\pm$ 0.3)} & 59.6 \tiny{($\pm$ 0.8)} & 30.0 \tiny{($\pm$ 0.3)} & 58.1 \tiny{($\pm$ 0.7)} & 5.5 \tiny{($\pm$ 0.9)} & {\bf 6.6} \tiny{($\pm$ 0.1)} & 43.6 \tiny{($\pm$ 0.5)} & 34.0 \tiny{($\pm$ 0.3)} \\

\rowcolor{lightgray} Osiris-R (Ours) & \underline{67.3} \tiny{($\pm$ 0.2)} & 4.9 \tiny{($\pm$ 0.2)} & 7.6 \tiny{($\pm$ 0.1)} & \underline{62.8} \tiny{($\pm$ 0.2)} & {\bf 32.3} \tiny{($\pm$ 0.1)} & \underline{59.9} \tiny{($\pm$ 0.2)} & 4.4 \tiny{($\pm$ 0.3)} & \underline{5.9} \tiny{($\pm$ 0.3)} & \underline{45.1} \tiny{($\pm$ 0.1)} & \underline{35.2} \tiny{($\pm$ 0.4)} \\

\rowcolor{lightgray} Osiris-D (Ours) & {\bf 67.8} \tiny{($\pm$ 0.1)} & 4.6 \tiny{($\pm$ 0.1)} & \underline{8.0} \tiny{($\pm$ 0.2)} & {\bf 64.2} \tiny{($\pm$ 0.1)} & \underline{32.0} \tiny{($\pm$ 0.1)} & {\bf 61.5} \tiny{($\pm$ 0.1)} & \underline{3.2} \tiny{($\pm$ 0.3)} & 5.6 \tiny{($\pm$ 0.5)} & {\bf 46.5} \tiny{($\pm$ 0.2)} & {\bf 35.6} \tiny{($\pm$ 0.1)} \\

\midrule

Offline & 69.2 \tiny{($\pm$ 0.3)} & - & - & 67.9 \tiny{($\pm$ 0.1)} & - & 60.7 \tiny{($\pm$ 0.2)} & - & - & 46.4 \tiny{($\pm$ 0.2)} & - \\

\bottomrule

\end{tabular}
}
\end{center}
\end{table}
\begin{table}[t]
\caption{Ablation of Osiris-D's components on 20-task Split-CIFAR-100. The ground truth task identity for each example is given to the model at test time, so the model predicts the most probable class within the task. $\mathcal{L}_\text{cross}$ benefits representations even in within-task discrimination when not considering classes from other tasks.}
\label{tab:wtl:ablation}
    \centering
    \resizebox{0.75\linewidth}{!}{
    \begin{tabular}{l|ccccc}
    \toprule
    &\bf A ($\uparrow$) & \bf F ($\downarrow$) & \bf K ($\uparrow$) & \bf C ($\uparrow$) & \bf T ($\uparrow$)  \\
    
    \midrule
    
    w/o isolated space & 85.6 \tiny{($\pm$ 0.3)} & 2.8 \tiny{($\pm$ 0.6)} & 4.9 \tiny{($\pm$ 0.2)} & 49.3 \tiny{($\pm$ 0.2)} & 29.8 \tiny{($\pm$ 0.1)} \\
    
    w/o $\mathcal{L}_\text{cross}$ & 84.4 \tiny{($\pm$ 0.2)} & 4.1 \tiny{($\pm$ 0.3)} & 
    {\bf 6.2} \tiny{($\pm$ 0.4)} & 47.2 \tiny{($\pm$ 0.2)} & 28.9 \tiny{($\pm$ 0.2)} \\
    
    w/o $\mathcal{L}_\text{past}$ & 86.3 \tiny{($\pm$ 0.3)} & 3.1 \tiny{($\pm$ 0.2)} & 6.1 \tiny{($\pm$ 0.3)} & 50.4 \tiny{($\pm$ 0.3)} & {\bf 30.0} \tiny{($\pm$ 0.1)} \\
    
    Full & {\bf 86.7} \tiny{($\pm$ 0.3)} & {\bf 2.2} \tiny{($\pm$ 0.3)} & 5.1 \tiny{($\pm$ 0.2)} & {\bf 51.3} \tiny{($\pm$ 0.1)} & {\bf 30.0} \tiny{($\pm$ 0.2)} \\
    
    \bottomrule
    
    \end{tabular}}
\end{table}%

We do not provide access to the task identity for each example at test time in Sec.~\ref{sec:exp}. In the scenario where task labels are given during evaluation~\citep{madaan2022representational}, it remains an open question whether consolidation provides benefits. We hypothesize that the consolidation term is beneficial because it potentially learns a broader set of features than FT by increasing the diversity of batches for the contrastive loss. 

To provide some evidence, we perform the same set of evaluations as in Sec.~\ref{sec:exp}, but with task labels given to the model. In other words, the model predicts the most probable class among classes within the same task as the ground truth class $i$ when calculating $A_{\cdot, i}$. The consolidation score (task-level KNN accuracy) is still calculated without task identities, and thus stays the same. The results are shown in Table~\ref{tab:wtl:random}, Table~\ref{tab:wtl:structured}, and Table~\ref{tab:wtl:ablation}. 
We find that the consolidation score still correlates with the accuracy, and \texttt{Osiris-D} consistently outperforms other models. In Table~\ref{tab:wtl:ablation}, without $\mathcal{L}_\mathrm{cross}$, \texttt{Osiris-D} experiences a 4.1\% consolidation score drop and a 2.3\% drop in accuracy. This shows that $\mathcal{L}_\mathrm{cross}$ benefits the representation even in within-task discrimination. 

Additionally, with consolidation, we expect the representations to separate all classes regardless of task identity. Therefore, they put the least assumptions on what classes we try to discriminate at test time and should appeal for a broader set of downstream use cases. For example, the test data is not required to be a subset of one of the tasks seen during training.

\section{Additional Plots}
\label{sec:appendix:plots}

\begin{figure}[!ht]
\begin{center}
\begin{subfigure}[t]{0.35\textwidth}
    \includegraphics[width=\linewidth]{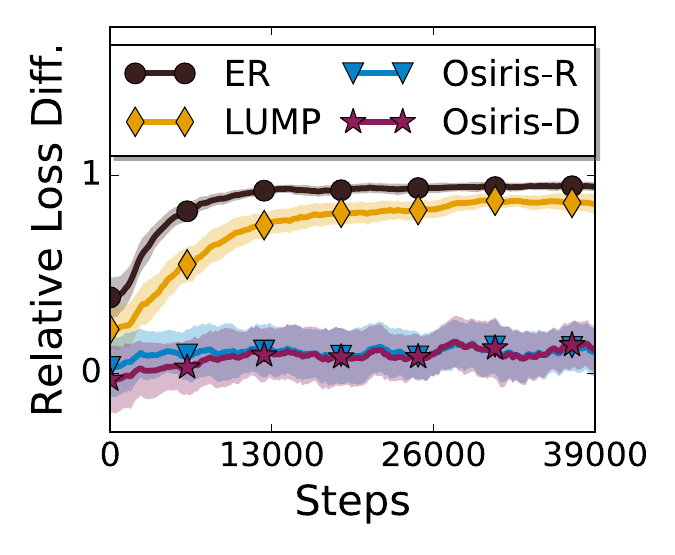}
    \caption{\texttt{Osiris}'s curves calculated with $g \circ f$.}
    \label{fig:appendix:overfit_curr}
\end{subfigure} %
\hspace{0.5in}
\begin{subfigure}[t]{0.35\textwidth}
    \includegraphics[width=\linewidth]{figures/overfitting_p2.pdf}
    \caption{\texttt{Osiris}'s curves calculated with $h \circ f$.}
    \label{fig:exp:overfit_past}
\end{subfigure} %
\end{center}
\caption{
Relative difference between contrastive loss on past-task data and on memory for replay-based methods. (a) All curves are calculated with the projector outputs where $\mathcal{L}_{\text{current}}$ is applied, i.e., with $g \circ f$. (b) Same as Fig.~\ref{fig:exp:overfit}, for \texttt{Osiris-D} and \texttt{Osiris-R}, we plot the curves calculated with the outputs of $h \circ f$ where $\mathcal{L}_{\text{past}}$ is applied. The curves for \texttt{ER} and \texttt{LUMP} are still calculated on their only projector branch, i.e, $g \circ f$. \texttt{Osiris-D} does not overfit on either branches. 
}
\label{fig:appendix:overfit}
\end{figure}

\begin{figure}[ht]
\begin{center}
\includegraphics[width=0.6\linewidth]{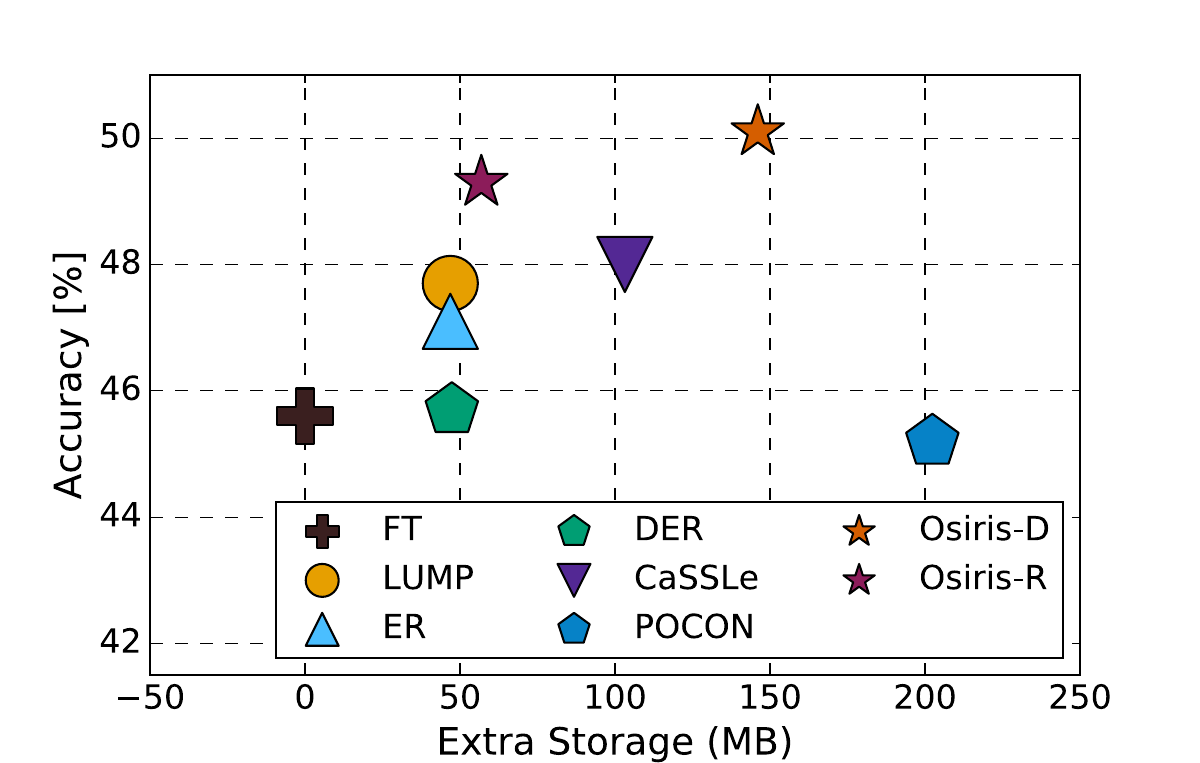}
\end{center}
\caption{Accuracy on 20-task Split-CIFAR-100 versus additional storage.
Storage is calculated by counting each additional model parameter besides the main model as well as each channel of each pixel in memory as a 64-bit float.} 
\label{fig:storage}
\end{figure}

\begin{figure}[t]
\centering
\makebox[0.03\textwidth]{}
\makebox[0.15\textwidth]{\textbf{Environment 1}}
\makebox[0.25\textwidth]{\textbf{Environment 2}}
\makebox[0.12\textwidth]{\textbf{Environment 3}}
\makebox[0.25\textwidth]{\textbf{Overall}}
\makebox[0.15\textwidth]{\textbf{Inter-Class Sim.}} 
\centering
\includegraphics[width=\linewidth]{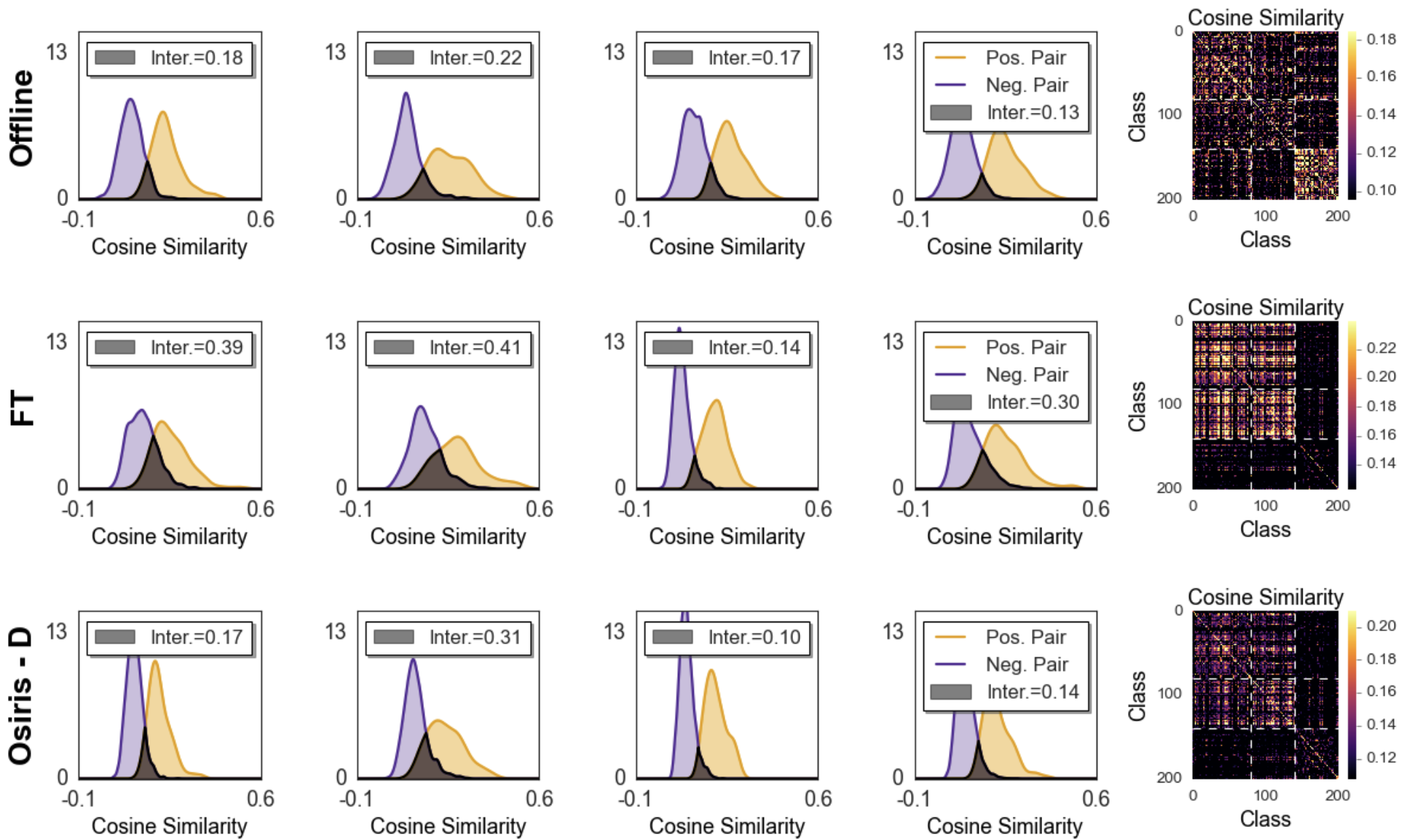}
\caption{Left: pairwise feature similarity distributions between examples from the same (positive) or different (negative) classes, within each environment of Structured Tiny-ImageNet, as well as on all data (overall). The densities are estimated with a Gaussian kernel. The intersections are marked with a darker shade and the values are obtained by integrating the shaded area. An empty intersection of supports of the two distributions sufficiently entails a KNN classifier with perfect accuracy, but it is not necessary. Right: mean cosine similarity between pairs of examples drawn from pairs of classes.}
\label{fig:sim_dist}
\end{figure}

\begin{figure}[!ht]
\begin{center}
    \includegraphics[width=\linewidth]{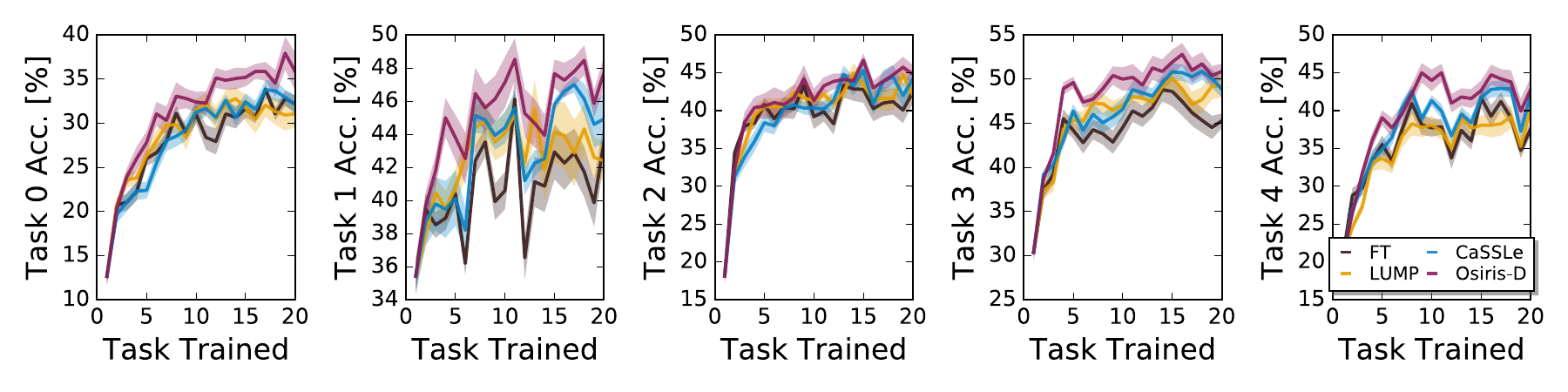}
\end{center}
\vspace{-0.15in}
\caption{Task 1-5 accuracy throughout training on 20-task Split-CIFAR-100.}
\label{fig:stability_0}
\end{figure}

\begin{figure}[!ht]
\begin{center}
    \includegraphics[width=\linewidth]{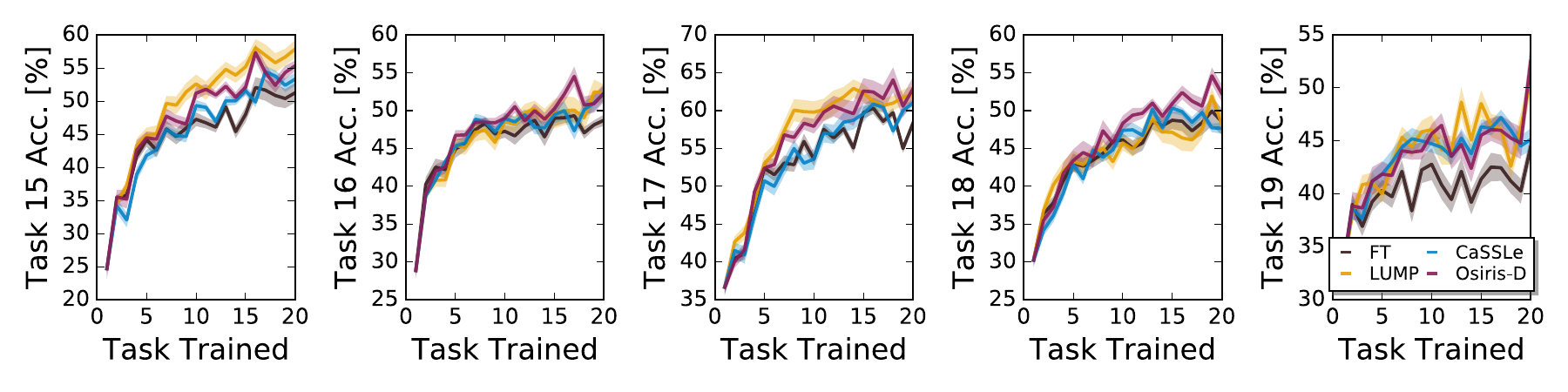}
\end{center}
\vspace{-0.15in}
\caption{Task 15-20 accuracy throughout training on 20-task Split-CIFAR-100.}
\label{fig:stability_3}
\end{figure}

In this section, we show four additional sets of plots. First, to complement Fig.~\ref{fig:exp:overfit}, we plot the relative loss difference curves for \texttt{Osiris-R} and \texttt{Osiris-D} on the outputs of $g \circ f$, i.e., on the space where $\mathcal{L}_{\text{current}}$ is applied, in Fig.~\ref{fig:appendix:overfit_curr}. Then, Fig.~\ref{fig:storage} shows the accuracy versus extra storage for different UCL models on 20-task Split-CIFAR-100. Fig.~\ref{fig:sim_dist} visualizes the pairwise cosine similarity distributions between examples from different class pairs. Finally, Fig.~\ref{fig:stability_0} through Fig.~\ref{fig:stability_3} plot the per-task accuracy of \texttt{FT}, \texttt{LUMP}, \texttt{CaSSLe}, and \texttt{Osiris-D}  throughout training on 20-task Split-CIFAR-100.

\clearpage

\begin{figure}[!ht]
\begin{center}
    \includegraphics[width=\linewidth]{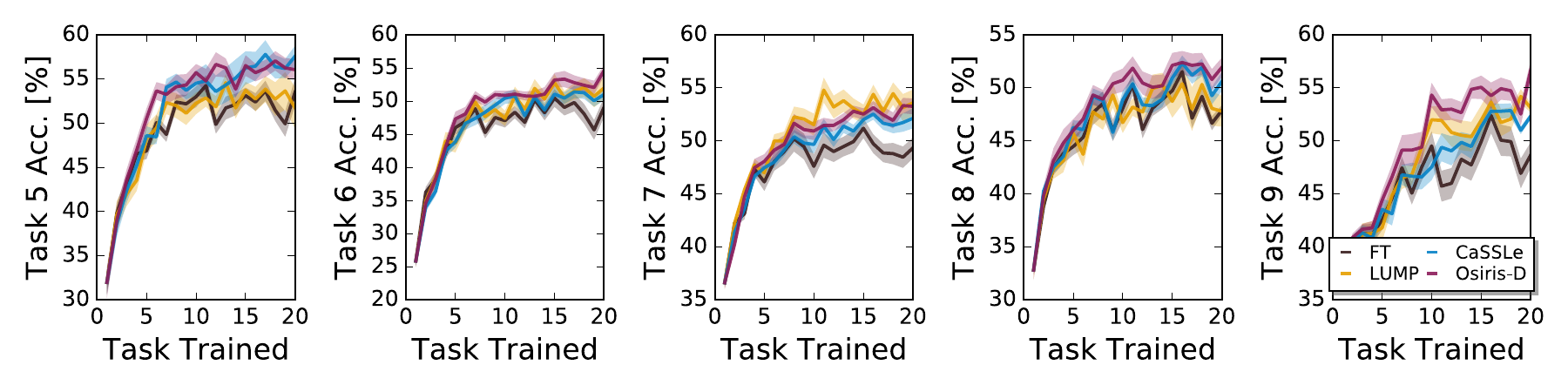}
\end{center}
\vspace{-0.15in}
\caption{Task 6-10 accuracy throughout training on 20-task Split-CIFAR-100.}
\end{figure}

\begin{figure}[!ht]
\begin{center}
    \includegraphics[width=\linewidth]{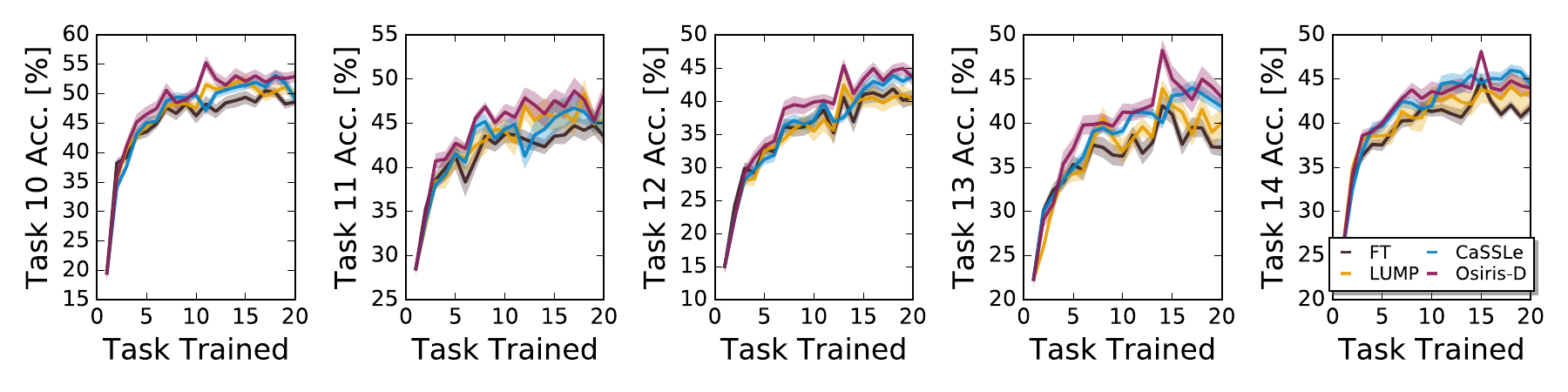}
\end{center}
\vspace{-0.15in}
\caption{Task 11-15 accuracy throughout training on 20-task Split-CIFAR-100.}
\end{figure}

\section{Structured Class Order}
\label{sec:appendix:order}

We list below the classes within each task in Structured Tiny-ImageNet which we described in Sec.~\ref{sec:exp}. Tasks 1-4 contain classes in an \texttt{indoor} environment. Task 5 contains classes in both \texttt{indoor} and \texttt{city} environments and serves as a soft transition. Tasks 6 and 7 contain classes in a \texttt{city} environment. Finally, tasks 9 and 10 contain classes in a natural, \texttt{wild} environment.

\begin{itemize}[leftmargin=*]
\item Task 1: trilobite, binoculars, American lobster, bow tie, volleyball, banana, fur coat, barbershop, sombrero, water jug, bathtub, beer bottle, bell pepper, hourglass, ice cream, altar, lampshade, boa constrictor, frying pan, Christmas stocking.

\item Task 2: turnstile, tabby, potter's wheel, chain, lemon, pill bottle, iPod, cockroach, oboe, punching bag, abacus, refrigerator, sock, bannister, candle, plate, ice lolly, Yorkshire terrier, apron, drumstick.

\item Task 3:
poncho, dining table, neck brace, guacamole, gasmask, backpack, academic gown, vestment, cash machine, CD player, espresso, potpie, syringe, orange, plunger, desk, Chihuahua, miniskirt, pretzel, bucket.

\item Task 4:
organ, chest, guinea pig, stopwatch, sandal, broom, pomegranate, barrel, wok, comic book, computer keyboard, meat loaf, pizza, basketball, remote control, teapot, mashed potato, teddy, cardigan, space heater.

\item Task 5:
Egyptian cat, rocking chair, wooden spoon, pop bottle, sunglasses, magnetic compass, sewing machine, jellyfish, beaker, Labrador retriever, dumbbell, nail, obelisk, lifeboat, steel arch bridge, moving van, gondola, military uniform, pole, beach wagon.

\item  Task 6:
freight car, torch, umbrella, rugby ball, limousine, projectile, brass, go-kart, confectionery, pay-phone, German shepherd, reel, trolleybus, crane, fountain, jinrikisha, convertible, tractor, butcher shop, thatch.

\item Task 7:
suspension bridge, bullet train, kimono, picket fence, water tower, school bus, maypole, birdhouse, sports car, beacon, parking meter, bikini, swimming trunks, flagpole, triumphal arch, cannon, Persian cat, scoreboard, police van, lawn mower.

\item Task 8:
dragonfly, scorpion, American alligator, tarantula, lion, golden retriever, mantis, bullfrog, African elephant, snail, bighorn, baboon, sea cucumber, brown bear, cougar, seashore, king penguin, koala, ladybug, tailed frog.

\item Task 9:
black widow, ox, grasshopper, acorn, fly, Arabian camel, coral reef, cliff dwelling, goldfish, goose, spider web, brain coral, barn, monarch, black stork, spiny lobster, standard poodle, sulphur butterfly, viaduct, albatross.

\item Task 10:
sea slug, chimpanzee, snorkel, slug, gazelle, dam, European fire salamander, hog, centipede, lesser panda, walking stick, lakeside, bee, mushroom, dugong, cauliflower, bison, alp, orangutan, cliff.
\end{itemize}

\end{document}